%% file: main.tex
\documentclass[10pt,twocolumn,letterpaper]{article}

\usepackage[pagenumbers]{cvpr} % To force page numbers, e.g. for an arXiv version
\input{preamble}

\definecolor{cvprblue}{rgb}{0.21,0.49,0.74}
\usepackage[pagebackref,breaklinks,colorlinks,citecolor=cvprblue]{hyperref}% 

\def\paperID{7480} % *** Enter the Paper ID herehttps://www.overleaf.com/project/650a371361c46f77c501c04e
\def\confName{CVPR}
\def\confYear{2024}

\begin{document}

\twocolumn[{%
\renewcommand\twocolumn[1][]{#1}%
\title{SwitchLight: Co-design of Physics-driven Architecture and \\ Pre-training Framework for Human Portrait Relighting}
\author{
Hoon Kim$^{1}$ \quad Minje Jang$^{1}$ \quad Wonjun Yoon$^{1}$ \quad Jisoo Lee$^{1}$ \quad Donghyun Na$^{1}$ \quad Sanghyun Woo$^{2}$ \\
\\
% \begin{center}
$^{1}$Beeble AI \hspace{0.5in} $^{2}$New York University
% \end{center}
}
\maketitle
% \vspace{-6mm}
\begin{center}
    \centering
    \includegraphics[width=\textwidth]{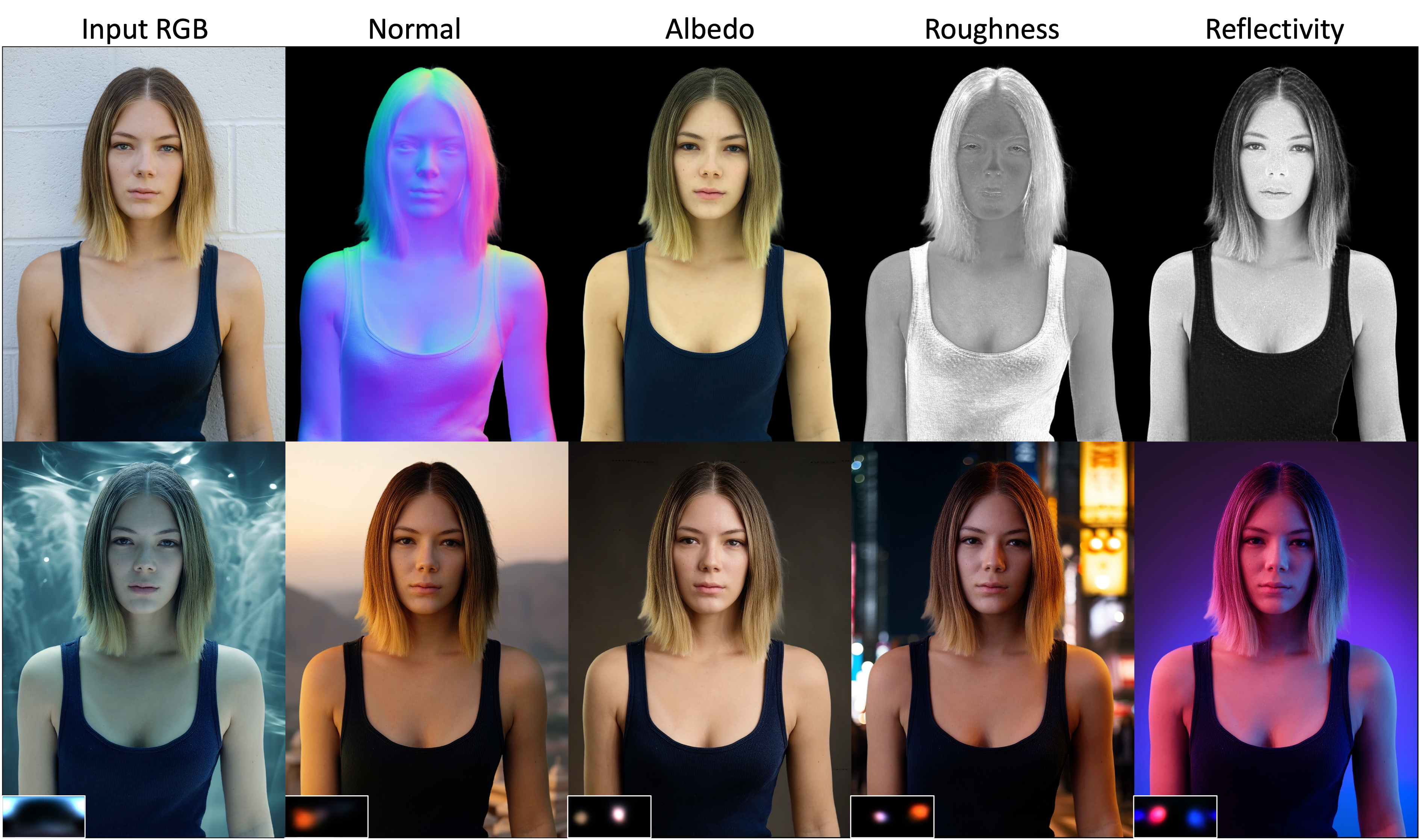}
    \vspace{-7mm}
    \captionof{figure}{
    \textbf{Be Anywhere at Any Time.}
    SwitchLight processes a human portrait by decomposing it into detailed intrinsic components,
    and re-renders the image under a designated target illumination, ensuring a seamless composition of the subject into any new environment.
    }
    \label{fig:teaser}
\end{center}%
}]

\maketitle
\def\thefootnote{*}\footnotetext{All authors contributed equally to this work}\def\thefootnote{\arabic{footnote}}
\input{sec/0_abstract}
\vspace{-12mm}
\input{sec/1_intro}

\input{sec/2_related_work}
\input{sec/3_architecture}
\input{sec/4_pretraining}
\input{sec/5_data}
\input{sec/6_experiments}

\input{sec/7_conclusion}
\input{sec/X_suppl}

\clearpage
{
    \small
    \bibliographystyle{ieeenat_fullname}
    \bibliography{main}
}


\end{document}

%% file: preamble.tex
\usepackage[dvipsnames]{xcolor}
\usepackage{tabularx}
\usepackage{amsmath}
\usepackage{graphicx}
\usepackage{booktabs}
\usepackage{multirow}
\usepackage{array}
\usepackage{adjustbox}
\usepackage{colortbl}

\newcommand{\gr}{\rowcolor[gray]{.95}}
\newcolumntype{g}{>{\columncolor[gray]{.95}}c}

\newlength\savewidth\newcommand\shline{\noalign{\global\savewidth\arrayrulewidth
  \global\arrayrulewidth 1pt}\hline\noalign{\global\arrayrulewidth\savewidth}}
\newcolumntype{x}[1]{>{\centering\arraybackslash}m{#1}}
\newcolumntype{y}[1]{>{\raggedright\arraybackslash}m{#1}}
\newcommand{\tablestyle}[2]{\setlength{\tabcolsep}{#1}\renewcommand{\arraystretch}{#2}\centering}

%% file: sec/0_abstract.tex
\begin{abstract}
\vspace{-1em}
We introduce a co-designed approach for human portrait relighting that combines a physics-guided architecture with a pre-training framework.
Drawing on the Cook-Torrance reflectance model, we have meticulously configured the architecture design to precisely simulate light-surface interactions.
Furthermore, to overcome the limitation of scarce high-quality lightstage data, we have developed a self-supervised pre-training strategy.
This novel combination of accurate physical modeling and expanded training dataset establishes a new benchmark in relighting realism.
\vspace{-1em}
\end{abstract}

%% file: sec/1_intro.tex
\section{Introduction}
% \label{sec:intro}
Relighting is more than an aesthetic tool; it unlocks infinite narrative possibilities and enables seamless integration of subjects into diverse environments (see Fig.~\ref{fig:teaser}).
This advancement resonates with our innate desire to transcend the physical constraints of space and time, while also providing tangible solutions to practical challenges in digital content creation.
It is particularly transformative in virtual (VR) and augmented reality (AR) applications, where relighting facilitates the real-time adaptation of lighting, ensuring that users and digital elements coexist naturally within any environment, offering a next level of telepresence.
% aenhancing both the realism and immersion.
%of the experience.

In this work, we focus on human portrait relighting.
While the relighting task fundamentally demands an in-depth understanding of geometry, material properties, and illumination, the challenge is more compounded when addressing human subjects, due to the unique characteristics of skin surfaces as well as the diverse textures and reflectance properties of a wide array of clothing, hairstyles, and accessories.
These elements interact in complex ways, necessitating advanced algorithms capable of simulating the subtle interplay of light with these varied surfaces.

Currently, the most promising approach involves the use of deep neural networks trained on pairs of high-quality relit portrait images and their corresponding intrinsic attributes, which are sourced from a light stage setup~\cite{debevec2000acquiring}.
Initial efforts approached the relighting process as a `black box'~\cite{sun2019single,wang2020single}, without delving into the underlying mechanisms.
Later advancements adopted a physics-guided model design, incorporating the explicit modeling of image intrinsics and image formation physics~\cite{nestmeyer2020learning}.
Pandey et al.~\cite{pandey2021total} proposed the Total Relight (TR) architecture, also physics-guided,
which decomposes an input image into surface normals and albedo maps, and performs relighting based on the Phong specular reflectance model.
The TR architecture has become foundational model for image relighting, with most recent and advanced architectures building upon its principle~\cite{yeh2022learning,ji2022geometry,mei2023lightpainter}.

Following the physics-guided approach, our contribution lies in a co-design of architecture with a self-supervised pre-training framework.
First, our architecture evolves towards a more accurate physical model by integrating the Cook-Torrance specular reflectance model~\cite{cook1982reflectance}, representing a notable advancement from the empirical Phong specular model~\cite{phong1998illumination} employed in the Total Relight architecture.
The Cook-Torrance model adeptly simulates light interactions with surface microfacets, accounting for spatially varying roughness and reflectivity. 
%This results in a more accurate and physically plausible rendering, essential for capturing the intricate details and unique reflectance characteristics of human skin and clothing textures.
Secondly, our pre-training framework scales the learning process beyond the typically hard-to-obtain lightstage data. By revisiting the masked autoencoder (MAE) framework~\cite{he2022masked}, we adept it for the task of relighting. These modifications are crafted to address the unique challenges posed by this task, enabling our model to learn from unlabelled data and refine its ability to produce realistic relit portraits during fine-tuning. 
To the best of our knowledge, this is the first time applying self-supervised pre-training specifically to the relighting task.
% To the best of our knowledge, this is the first time we have designed and applied self-supervised pre-training to relighting tasks.

To summarize, our contribution is twofold. Firstly, by enhancing the physical reflectance model, we have introduced a new level of realism in the output. Secondly, by adopting self-supervised learning, we have expanded the scale of the training data and enhanced the expression of lighting in diverse real-world scenarios. Collectively, these advancements have led SwitchLight framework to achieve a new state-of-the-art in human portrait relighting.

%% file: sec/2_related_work.tex
\section{Related Work}

\noindent\textbf{Human Portrait Relighting} is an ill-posed problem due to its under-constrained nature.
To tackle this, earlier methods incorporated 3D facial priors~\cite{shu2017portrait}, exploited image intrinsics~\cite{barron2014shape,sengupta2018sfsnet}, or framed the task as a style transfer~\cite{shih2014style}.
Light stage techniques~\cite{wenger2005performance} offer a more powerful solution  by recording subject's reflectance fields under varying lighting conditions~\cite{dorsey1995interactive,debevec2000acquiring}, though they are labor-intensive and require specialized equipment.
A promising alternative has emerged with deep learning, utilizing neural networks trained on light stage data.
Sun et al.~\cite{sun2019single} pioneered this approach, but their method had limitations in representing non-Lambertian effects. %due to its insufficient adherence to physical laws. 
This was improved upon by Nestmeyer et al.~\cite{nestmeyer2020learning}, who integrated rendering physics into network design, albeit limited to directional light. Building upon this, Pandey et al.~\cite{pandey2021total} incorporated the Phong reflection model and a high dynamic range (HDR) lighting map~\cite{debevec2008rendering} into their network, enabling a more accurate representation of global illumination.
Simultaneously, efforts have been made to explore portrait relighting without light stage data~\cite{zhou2019deep,sengupta2021light,hou2021towards,hou2022face,wang2023sunstage}.
Moreover, introduction of NeRF~\cite{chen2022relighting4d} and diffusion-based~\cite{ponglertnapakorn2023difareli} models has opened new avenues in the field.
However, networks trained with light-stage data maintain superior accuracy and realism, thanks to physics-based composited relight image training pairs and precise ground truth image intrinsics~\cite{zhou2023relightable}.

Our work furthers this domain by integrating the Cook-Torrance model into our network design, shifting from the empirical Phong model to a more physics-based approach, thereby enhancing the realism and detail in relit images.

\vspace{0.5em}
\noindent\textbf{Self-supervised Pre-training} has become a standard training scheme in the development of large language models like BERT~\cite{devlin2018bert} and GPT~\cite{radford2018improving}, and is increasingly influential in vision models, aiming to replicate the `BERT moment'.
This approach typically involves pre-training on extensive unlabeled data, followed by fine-tuning on specific tasks.
While early efforts in vision models focused on simple pre-text tasks~\cite{doersch2015unsupervised,noroozi2016unsupervised,pathak2016context,zhang2016colorful,gidaris2018unsupervised}, the field has evolved through stages like contrastive learning~\cite{chen2020simple,he2020momentum} and masked image modeling~\cite{bao2021beit,he2022masked,woo2023convnext}. However, the primary focus has remained on visual recognition, with less attention to other domains. Exceptions include low-level image processing tasks~\cite{chen2021pre,li2021efficient,chen2023activating,liu2023degae} using the vision transformer~\cite{dosovitskiy2020image}.

Our research takes a different route, focusing on human portrait relighting—a complex challenge of manipulating illumination in the image. This direction is crucial because acquiring accurate ground truth data, especially from light stage, is both expensive and difficult. We modify the MAE framework~\cite{he2022masked}, previously successful in robust image representation learning and developing locality biases~\cite{park2023self}, to suit the unique requirements of effective relighting. 

%% file: sec/3_architecture.tex
\section{SwitchLight}

\begin{figure*}[!t]
  \includegraphics[width=\linewidth]
  {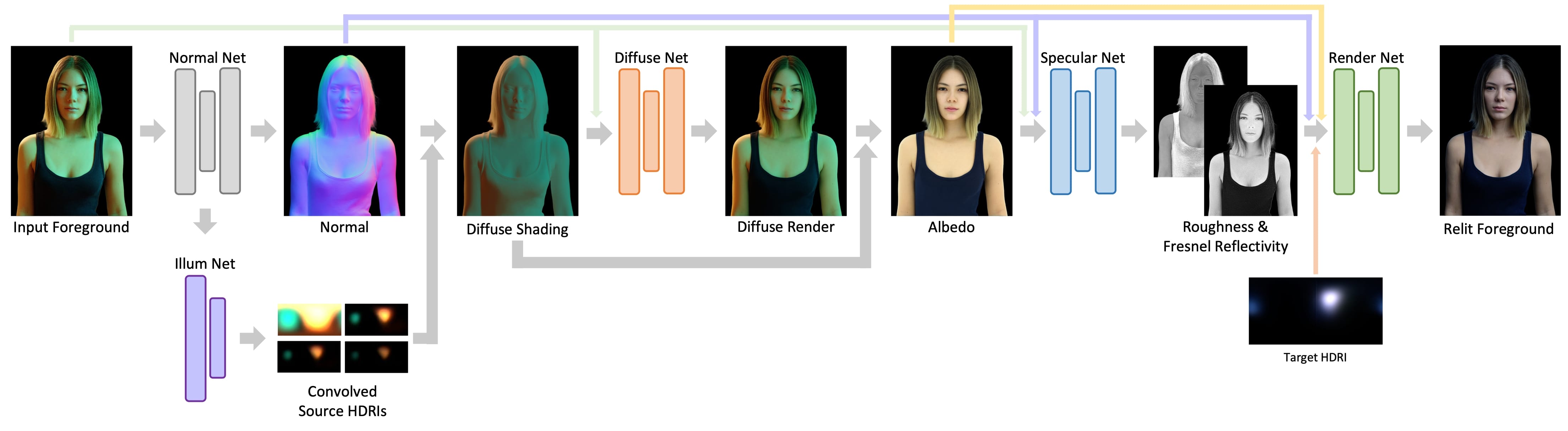}
  % {./figures/fig2_architecture_v4_a.png}
  % {./figures/fig2_architecture_v4_full.png}
  \vspace{-6mm}
  % \captionsetup{font=footnotesize}
  \caption{\textbf{SwitchLight Architecture.}
  The input source image is decomposed into \emph{normal} map, \emph{lighting}, \emph{diffuse} and \emph{specular} components. Given these intrinsics, images are re-rendered under target lighting.
  The architecture integrates the \emph{Cook-Torrance} reflection model; the final output combines physically-based predictions with neural network enhancements for realistic portrait relighting.
  }
  \label{fig:switch_light}
  \vspace{-2mm}
\end{figure*}

We introduce SwitchLight, a state-of-the-art framework for human portrait relighting, with its architectural overview presented in Fig.~\ref{fig:switch_light}.
We first provide foundational concepts in Sec.~\ref{sec:preliminaries},
and define the problem in Sec.~\ref{sec:problem_formulation}.
This is followed by detailing the architecture in Sec.~\ref{sec:architecture}, and lastly, we describe the loss functions used in Sec.~\ref{sec:objectives}.
% \begin{table}
%     \centering
%     \begin{tabularx}{\linewidth}{|c|X|}
%         \hline
%         \textbf{Symbol} & \textbf{Definition} \\
%         \hline
%         \( v \) & View unit vector \\
%         \( l \) & Incident light unit vector \\
%         \( n \) & Surface normal unit vector \\
%         \( h \) & Half unit vector between \( l \) and \( v \) \\
%         \( f \) & Bidirectional Reflectance Distribution Function (BRDF) \\
%         \( f_d \) & Diffuse component of a BRDF \\
%         \( f_s \) & Specular component of a BRDF \\
%         \( \alpha \) & Roughness \\
%         \( \sigma \) & Albedo \\
%         \( \Omega \) & Spherical domain \\
%         \( f_0 \) & Reflectance at normal incidence \\
%         \( k_s \) & Phong specular coefficient \\
%         \( p \) & Phong exponent \\
%         \( r \) & Reflected light vector\\
%         \( \langle \cdot \rangle \) & Dot product clamped to [0..1] \\
%         \hlinehttps://www.overleaf.com/project/650a371361c46f77c501c04e
%     \end{tabularx}
%     \caption{Symbols definitions}
% \end{table}

\subsection{Preliminaries}
\label{sec:preliminaries}
In this section, vectors $\mathbf{n}$, $\mathbf{v}$, $\mathbf{l}$, and $\mathbf{h}$ are denoted as unit vectors. Specifically, $\mathbf{n}$ represents the surface normal, $\mathbf{v}$ is the view direction, $\mathbf{l}$ is the incident light direction, and $\mathbf{h}$ is the half-vector computed from $\mathbf{l}$ and $\mathbf{v}$. The dot product is clamped between $[0..1]$, indicated by $\langle \cdot \rangle$.

\paragraph{Image Rendering.}
The primary goal of image rendering is to create a visual representation that accurately simulates the interactions between light and surfaces. These complex interactions are encapsulated by the rendering equation:
\begin{equation}
    L_o(\mathbf{v}) = \int_{\Omega} f(\mathbf{v}, \mathbf{l}) L_i(\mathbf{l}) \langle \mathbf{n} \cdot \mathbf{l} \rangle \, d\mathbf{l}
    \label{eq:rendering}
\end{equation}
where $L_o(\mathbf{v})$ denotes the radiance, or the light intensity perceived by the observer in direction $\mathbf{v}$. 
It is the cumulative result of incident lights $L_i(\mathbf{l})$ from all possible directions over the hemisphere, $\Omega$, centered around the surface \emph{normal}, denoted as $\mathbf{n}$.
Central to this equation lies the Bidirectional Reflectance Distribution Function (BRDF), denoted as $f(\mathbf{v},\mathbf{l})$, describing the surface's reflection characteristics.

\paragraph{BRDF Composition.}
The BRDF, represented by $f(\mathbf{v},\mathbf{l})$, describes how light is reflected at an opaque surface. It is composed of two major components: diffuse reflection ($f_d$) and specular reflection ($f_s$):
\begin{equation}
    f(\mathbf{v}, \mathbf{l}) = f_d(\mathbf{v}, \mathbf{l}) + f_s(\mathbf{v}, \mathbf{l})
    \label{eq:brdf_composition}
\end{equation}
A surface intrinsically exhibits both diffuse and specular reflections.
The diffuse component uniformly scatters light, ensuring consistent illumination regardless of the viewing angle.
In contrast, the specular component is viewing angle-dependent, producing shiny highlights that are crucial for achieving a photorealistic effect.

\paragraph{Lambertian Diffuse Reflectance.}
Lambertian reflectance, a standard model for diffuse reflection,
describes a uniform light scatter irrespective of the viewing angle. 
This ensures a consistent appearance in brightness:
\begin{equation}
    f_d(\mathbf{v}, \mathbf{l}) = \frac{\sigma}{\pi} \hspace{.5em} \text{[const.]}
    \label{eq:lambertian_diffuse}
\end{equation}
Here, $\sigma$ is the \emph{albedo}, indicating the intrinsic color and brightness of the surface.

\paragraph{Cook-Torrance Specular Reflectance.}
The Cook-Torrance model, based on microfacet theory, represents surfaces as a myriad of tiny, mirror-like facets.
It incorporates a \textit{roughness} parameter $\alpha$, which allows precise rendering of surface specular reflectance:
\begin{equation}
    f_{s}(\mathbf{v}, \mathbf{l}) = \frac{D(\mathbf{h}, \alpha)G(\mathbf{v}, \mathbf{l}, \alpha)F(\mathbf{v}, \mathbf{h}, f_0)}{4 \langle \mathbf{n} \cdot \mathbf{l} \rangle \langle \mathbf{n} \cdot \mathbf{v} \rangle}
    \label{eq:cook_torrance}
\end{equation}
In this model, 
$D$ is the microfacet distribution function, describing the orientation of the microfacets relative to the half-vector $h$,
$G$ is the geometric attenuation factor, accounting for the shadowing and masking of microfacets,
and $F$ is the Fresnel term, calculating the reflectance variation depending on the viewing angle, where \( f_0 \) is the surface \emph{Fresnel reflectivity} at normal incidence.
A lower \( \alpha \) value implies a smoother surface with sharper specular highlights, whereas a higher \( \alpha \) value indicates a rougher surface, resulting in more diffused reflections. By adjusting \( \alpha \), the Cook-Torrance model can depict a range of specular reflections.
%, from glossy to matte.

\paragraph{Image Formation.}
Upon the base rendering equation, we include the diffuse and specular components of the BRDF and derive a unified formula:
\begin{equation}
L_o(\mathbf{v}) = \int_{\Omega} \left(f_d(\mathbf{v}, \mathbf{l}) + f_s(\mathbf{v}, \mathbf{l})\right) E(\mathbf{l}) \langle \mathbf{n} \cdot \mathbf{l} \rangle \, d\mathbf{l}
\label{eq:gen_rendering}
\end{equation}
where \( E(\mathbf{l}) \) denotes the incident environmental lighting.
This formula represents the core principle that an image is a product of interplay between the BRDF and lighting.
To further clarify this concept, we introduce a rendering function \( R \), succinctly modeling the process of image formation:
\begin{equation}
I = R({\color{blue}\underbrace{\color{black}\mathbf{n}, \sigma, \alpha, f_0}_{\text{surface attributes}}}, {\color{blue}\underbrace{\color{black}E\vphantom{\mathbf{n}, \sigma, \alpha, f_0}}_{\text{lighting}}})
\label{eq:formation}
\end{equation}
It is important to note that since the BRDF is a function of surface properties, as detailed in Eqn.~\ref{eq:lambertian_diffuse} and \ref{eq:cook_torrance}, we can now clearly understand that image formation is essentially governed by the interaction of surface attributes and lighting.

% $L_i(\mathbf{l})$ is the irradiance, representing the incident light on the surface from direction $\mathbf{l}$. 
%The term $\langle \mathbf{n} \cdot \mathbf{l}\rangle$ implements the cosine law, adjusting for the angle between the normal $\mathbf{n}$ and light direction $\mathbf{l}$. Light aligned with the normal contributes the most, whereas light from oblique angles contributes less. 
% \paragraph{Phong Specular Reflectance.}
% The Phong reflection model is an empirical method for approximating reflection on glossy surfaces, which is defined as:
% \begin{equation}
%     f_{s, \text{phong}}(\mathbf{v}, \mathbf{l}) = k_s \langle \mathbf{v} \cdot \mathbf{r} \rangle^p
%     \label{eq:phong_specular}
% \end{equation}
% It describes the reflection on shiny surfaces by capturing how closely the reflected light, $\mathbf{r} = 2(\mathbf{n} \cdot \mathbf{l})\mathbf{n} - \mathbf{l}$, aligns with the viewer's perspective, $\mathbf{v}$, accentuated by its shiness factor $p$. The specular coefficient, $k_s$, modulates the intensifty of this reflection.
% While the Phong model is versatile and visually effective, it lacks a firm basis in real-world physics.

\subsection{Problem Formulation}
\label{sec:problem_formulation}
\paragraph{Image Relighting.}
Given the image formation model above, our goal is to manipulate the lighting of an existing image.
This involves two main steps: inverse rendering and re-rendering under target illumination, both driven by neural networks. For a given source image \( I_\text{src} \) and target illumination \( E_\text{tgt} \), the process is delineated as:
% \minje{set to vec, D to ?}
\begin{align*}
\text{Inverse Rendering.} \hspace{.5em} &(\mathbf{n}, \sigma, \alpha, f_0, E_\text{src}) = U(I_\text{src})   \\
\text{Rendering with Target Light.} \hspace{.5em} &I_\text{tgt} = R(\mathbf{n}, \sigma, \alpha, f_0, E_\text{tgt})
\end{align*}
During the inverse rendering step, the function $U$ unravels the intrinsic properties of \( I_\text{src} \).
In the subsequent relighting step, the derived intrinsic properties along with new illumination \( E_\text{tgt} \) are employed by the rendering function $R$ to generate the target relit image \( I_\text{tgt} \). 
% This function not only adheres to the rendering equation but also captures the `residual' complexities, those aspects of the scene appearance that are not fully captured by thehttps://www.overleaf.com/project/650a371361c46f77c501c04e physical models.
%and produces the set \(\{\mathbf{n}, \sigma, \alpha, f_0, E_s\}\).
% The function is trained to confrom to the image formation process described in Eqn.~\ref{eq:gen_rendering}, with its specular model following the Cook-Torrance reflection model.

\subsection{Architecture}
\label{sec:architecture}
Our architecture systematically
executes the two primary stages outlined in our problem formulation.
% We utilize UNet-based~\cite{ronneberger2015u} neural networks, as illustrated in Fig.~\ref{fig:switch_light}.
The first stage involves extracting intrinsic properties from the source image \( I_\text{src} \).
For this purpose, we employ a matting network~\cite{sengupta2020background,lin2021real,kim2022revisiting} to accurately separate the foreground. This extracted image is then processed by our inverse rendering network \(U\), which infers normal, albedo, roughness, reflectivity, and source lighting.
Subsequently, the second stage involves re-rendering the image under new target lighting conditions.
To achieve this, the acquired intrinsics, along with the target lighting \( E_\text{tgt} \), are fed into our relighting network \(R\), producing the relit image \( I_\text{tgt} \).

% Four of these networks perform inverse rendering tasks, decomposing the image into its normal map, source illumination, diffuse reflectance and specular components.
% The final network re-renders the image by synthesizing these intrinsic elements under the new target illumination.

% \woo{we need an architecture table, maybe in supp}

\paragraph{\textit{Normal Net.}}
The network takes the source image \( I_\text{src} \) $\in$ \( \mathbb{R}^{\text{H} \times \text{W} \times 3} \) and generates a \textbf{normal map} \(\mathbf{\hat{N}}\). Each pixel in this map contains a unit normal vector \(\mathbf{\hat{n}}\), indicating the orientation of the corresponding surface point.

\paragraph{\textit{Illum Net.}}
The network infers the lighting conditions in the given image captured in an HDRI format.
Specifically, it computes the \textbf{convolved HDRIs}: %which are derived using predefined Phong reflectance lobes:
\begin{equation}
E^{p}_{\text{src}}(\mathbf{l'}) = \int_{\Omega} \color{blue}{\underbrace{\color{black}{E_\text{src}(\mathbf{l})}}_{\text{HDRI}}} \, \color{blue}{\underbrace{\color{black}{\langle \mathbf{l'} \cdot \mathbf{l} \rangle^p}}_{\text{Phong lobe}}} \, \color{black}{d\mathbf{l}}
\label{eq:phong_lobe}
\end{equation}
% \minje{In this equation, \(E_\text{src}\) $\in$ \( \mathbb{R}^{\text{\# of directional lights} \times 3}\) is the original source illumination, with \(\mathbf{l}\) indicating spherical directions of each directional light.
% The term \(\langle \mathbf{l'} \cdot \mathbf{l} \rangle^p\) represents the Phong reflectance lobe with shininess exponents \(p \in \{1, 16, 32, 64\}\), which incorporates various specular terms. Consequently, it is expressed in a multi-dimensional tensor form as \(\mathbb{R}^{4 \times \text{H}_\text{cHDRI} \times \text{W}_\text{cHDRI} \times \text{H}_\text{HDRI} \times \text{W}_\text{HDRI}}\).
% Finally, \(E^{p}_{\text{src}}\) $\in$ \( \mathbb{R}^{4 \times \text{H}_\text{cHDRI} \times \text{W}_\text{cHDRI} \times 3}\) is the convolved HDRI.
% %, and the network is tasked to infer this lighting representation.
% In this work, we use the resolution of HDRI and convolved HDRI as 32 \(\times\) 64 and 64 \(\times\) 128, respectively.}
In this equation, \(E_\text{src}\) $\in$ \( \mathbb{R}^{\text{H}_\text{HDRI} \times \text{W}_\text{HDRI} \times 3}\) is the original source HDRI map,  with \(\mathbf{l}\) indicating spherical directions in the HDRI space \(\mathbb{R}^{\text{H}_\text{HDRI} \times \text{W}_\text{HDRI}}\).
The term \(\langle \mathbf{l'} \cdot \mathbf{l} \rangle^p\) represents the Phong reflectance lobe with shininess exponents \(p \in \{1, 16, 32, 64\}\), which incorporates various specular terms. Consequently, it is expressed in a multi-dimensional tensor form as \(\mathbb{R}^{4 \times \text{H}_\text{cHDRI} \times \text{W}_\text{cHDRI} \times \text{H}_\text{HDRI} \times \text{W}_\text{HDRI}}\).
Finally, \(E^{p}_{\text{src}}\) $\in$ \( \mathbb{R}^{4 \times \text{H}_\text{cHDRI} \times \text{W}_\text{cHDRI} \times 3}\) is the convolved HDRI.
%, and the network is tasked to infer this lighting representation.
In this work, we set the resolution of HDRI and convolved HDRI at 32 \(\times\) 64 and 64 \(\times\) 128, respectively, and we apply convolution on light source coordinates.

% \(\text{H}_{\text{HDRI}} \times \text{W}_{\text{HDRI}} = 32 \times 64\) and \(\text{H}_{\text{cHDRI}} \times \text{W}_{\text{cHDRI}} = 64 \times 128\).

The network employs a cross-attention mechanism at its core, where predefined Phong reflectance lobes serve as queries, and the original image acts as both keys and values.
Within this setup, the convolved HDRI maps are synthesized by integrating image information into the Phong reflectance lobe representation.
Specifically, our model utilizes bottleneck features from the \textit{Normal Net} as a compact image representation.
Our approach simplifies the complex task of HDRI reconstruction by instead focusing on estimating interactions with known surface reflective properties.
% \woo{experiment needed: HDRI vs ConvHDRI}
% \woo{brief note about video relighting}

\paragraph{\textit{Diffuse Net.}}
Estimating albedo is challenging due to the ambiguities in surface color and material properties, further complicated by shadow effetcs.
To address this, we prioritize the inference of source \textbf{diffuse render}, \(I_{\text{src},\text{diff}}\):
\begin{equation}
L_{\text{src},o_\text{diff}}(\mathbf{v}) = {\color{blue}\underbrace{\color{black}\frac{\sigma}{\pi} \vphantom{\int_{\Omega} E(\mathbf{l})}}_{\text{diffuse BRDF}}} {\color{blue}\underbrace{\color{black}\int_{\Omega} E_\text{src}(\mathbf{l}) \langle \mathbf{n} \cdot \mathbf{l} \rangle \, d\mathbf{l}}_{\text{diffuse shading}}}
\label{eq:diffuse_render}
\end{equation}
Our key insight is that the diffuse render closely resembles the original image, which simplifies the model learning process. It captures surface color after removing specular reflections, such as shine or gloss, contrasting with albedo that represents the true surface color unaffected by lighting and shadows.
The network takes a source image  \( I_\text{src} \), concatenated with its diffuse shading, to produce the diffuse render.
As in Eqn.~\ref{eq:phong_lobe}, the diffuse shading, \( \hat{E}^{1}_{\text{src}} (\hat{\mathbf{n}}) \), is derived using the predicted normals, \(\hat{\mathbf{n}}\), and the predicted lighting map, \( \hat{E}^{1}_{\text{src}}\), with a Phong exponent of 1 for the diffuse term.
The \textbf{albedo map} \(\mathbf{\hat{A}}\) is then computed by dividing the predicted diffuse render by its diffuse shading:
\begin{equation}
\frac{\hat{\sigma}}{\pi} = \frac{\hat{L}_{\text{src},o_\text{diff}}(\mathbf{v})}{\hat{E}^{1}_{\text{src}}(\hat{\mathbf{n}})}
\label{eq:ours_albedo}
\end{equation}
We have empirically validated that it significantly enhances albedo prediction across a range of real-world scenarios.

\paragraph{\textit{Specular Net.}}
The network infers surface attributes associated with the Cook-Torrance specular elements, specifically, the \textbf{roughness} \( \alpha \) and \textbf{Fresnel reflectivity} \( f_0 \).
It uses a source image, predicted normal, and albedo maps as inputs.

\begin{figure}[!t]
  \includegraphics[width=\linewidth]
  {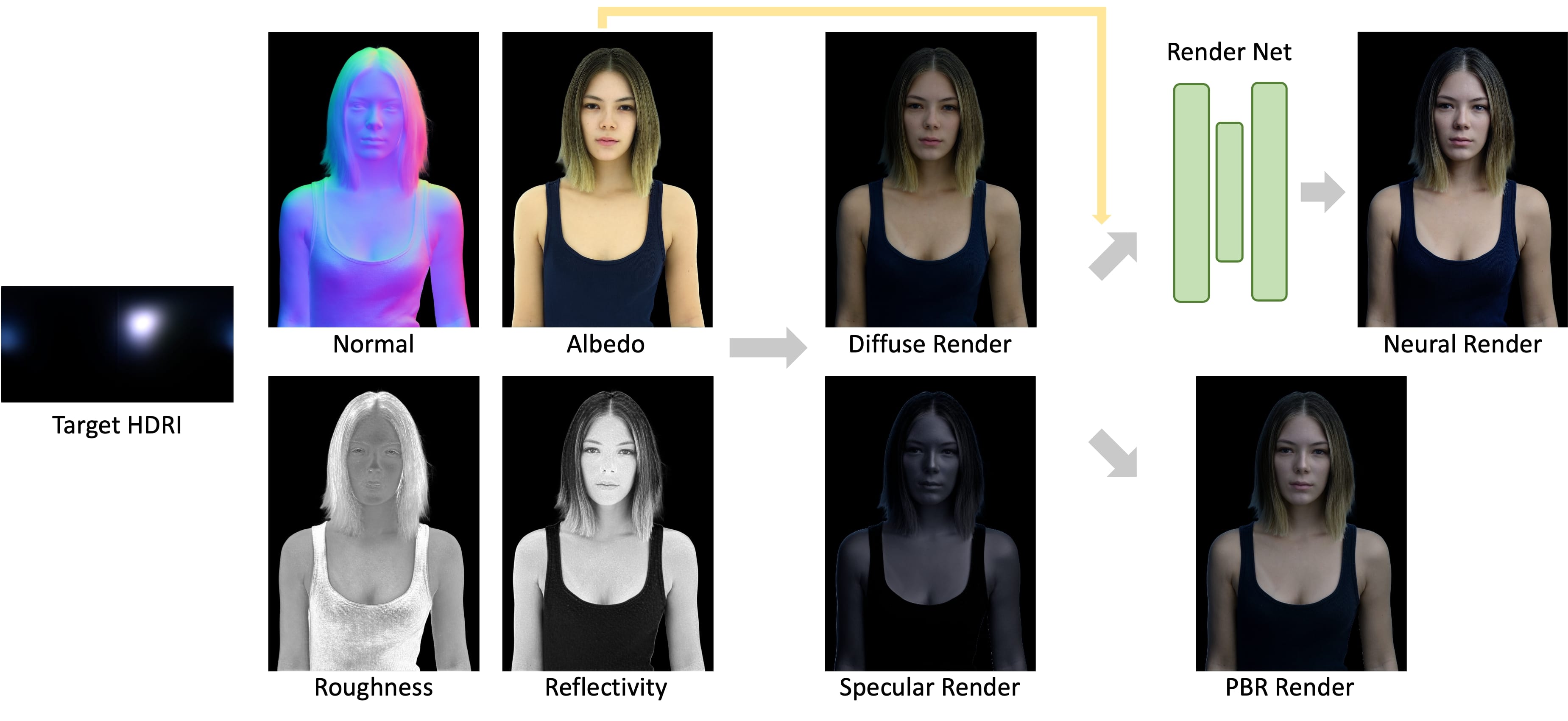}
  % {./figures/fig2_architecture_v5_b.png}
  % {./figures/fig2_architecture_v4_full.png}
  \vspace{-6mm}
  % \captionsetup{font=footnotesize}
  \caption{\textbf{Render Net Overview.}
  Utilizing extracted image intrinsics, it employs the Cook-Torrance model for initial relighting and a neural network for enhanced refinement, producing high-fidelity relit images through a synergistic computational approach.
  }
  \label{fig:render_net}
  \vspace{-3mm}
\end{figure}

\begin{figure}[!t]
  \includegraphics[width=\linewidth]{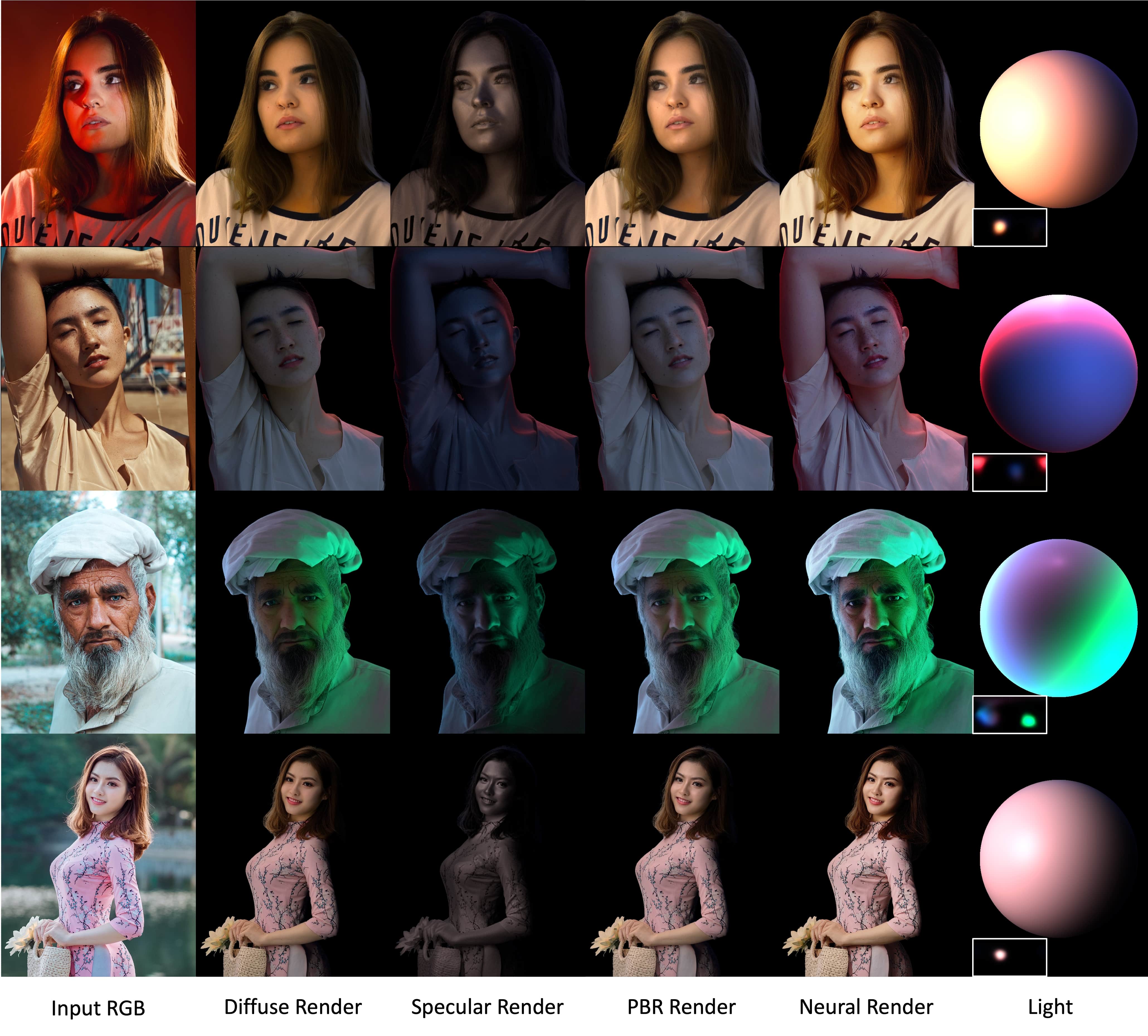}
  \vspace{-6mm}
  % \captionsetup{font=footnotesize}
  \caption{\textbf{Neural Render Enhancement.} 
  Using the Cook-Torrance model, diffuse and specular renders are computed, which are then composited into a physically-based rendering. Subsequently, a neural network enhances this PBR render, improving aspects such as brightness and specular details.
  }
  \label{fig:renders}
  \vspace{-3mm}
\end{figure}

\paragraph{\textit{Render Net.}}
The network utilizes extracted intrinsic surface attributes to produce the \textbf{target relit images}.
It generates two types of relit images, as shown in Fig.~\ref{fig:render_net}.
The first type adheres to the physically-based rendering (PBR) principles of the Cook-Torrance model. This involves computing diffuse and specular renders under the target illumination using Eqn.~\ref{eq:lambertian_diffuse} and Eqn.~\ref{eq:cook_torrance}. These renders are combined to form the PBR render, \( \hat{I}^\text{PBR}_\text{tgt} \), as:
\begin{equation}
\hat{L}^{\text{PBR}}_{\text{tgt},o}(\mathbf{v}) = \hat{L}_{\text{tgt},o_\text{diff}}(\mathbf{v}) + \hat{L}_{\text{tgt},o_\text{spec}}(\mathbf{v})
\label{eq:pred_relit}
\end{equation}
The second type of relit image is the result of a neural network process.
This enhances the PBR render, capturing finer details that the Cook-Torrance model might miss.
It employs the albedo, along with the diffuse and specular renders from the Cook-Torrance model, to infer a more refined target relit image, termed the neural render, \( \hat{I}^\text{Neural}_\text{tgt} \).
The qualitative improvements achieved through this neural enhancement are illustrated in Fig.~\ref{fig:renders}.

% For both output types, reconstruction losses are applied to ensure that the intrinsic components align with the physical rendering equation and that the network effectively learns to render the image with high fidelity.

\subsection{Objectives}
\label{sec:objectives}
We supervise both intrinsic image attributes and relit images using their corresponding ground truths, obtained from the lightstage.
% The attributes include the normal map, convolved HDRI maps, diffuse render, and albedo map. 
% Additionally, the roughness and Fresnel reflectivity are implicitly learned by applying reconstruction loss to the PBR render.
We employ a combination of reconstruction, perceptual~\cite{johnson2016perceptual}, adversarial~\cite{isola2017image}, and specular~\cite{pandey2021total} losses.

\paragraph{Reconstruction Loss \( \mathcal{L} = \ell_1(\text{X}, \hat{\text{X}}) \).} 
It measures the pixel-level differences between the ground truth \(\text{X}\) and its prediction \(\hat{\text{X}}\).
This loss is applied across different attributes, including normal map \(\mathbf{\hat{N}}\), convolved source HDRI maps \(\hat{E}^{p}_\text{src}\), source diffuse render \(\hat{I}_{\text{src},\text{diff}}\), albedo map  \(\mathbf{\hat{A}}\), and both types of target relit images \(\hat{I}^{\text{PBR}}_\text{tgt}\) and \( \hat{I}^{\text{Neural}}_\text{tgt}\). 
The attributes like the roughness \(\alpha\) and Fresnel reflectivity \(f_0\) are also implicitly learned when this loss applied to the PBR render.

\paragraph{Perceptual Loss \(\mathcal{L}_{\text{vgg}} = \ell_2(\text{VGG}(\text{X}), \text{VGG}(\hat{\text{X}}))\).} 
It captures high-level feature differences based on a VGG-network feature comparison. We apply this loss to the source diffuse render, albedo, and target relit images.
%to improve their perceptual similarity.

\paragraph{Adversarial Loss \(\mathcal{L}_{\text{adv}} = \text{GAN}(\text{X}, \hat{\text{X}})\).}
It promotes realism in the generated images by fooling a discriminator network. This loss is applied to the target relit images. 
We employ a PatchGAN architecture, with detailed specifications provided in the supplementary material.

\paragraph{Specular Loss \(
    \mathcal{L}_{\text{spec}} = \ell_1(\text{X} \odot \hat{\text{S}}, \hat{\text{X}} \odot \hat{\text{S}})
\).} 
It enhances the specular highlights in the relit images.
Specifically, we utilize the predicted specular render \( \hat{\text{S}} := \hat{I}^{\text{PBR}}_{\text{tgt},\text{spec}} \) derived from the Cook-Torrance physical model, to weigh the \(\ell_1\) reconstruction loss.
Here, \( \odot \) denotes the element-wise multiplication. 
This loss is applied to the neural render.

\paragraph{Final Loss.} 
The SwitchLight is trained in an end-to-end manner using the weighted sum of the above losses:
\begin{equation}
\begin{aligned}
\mathcal{L}_{\text{relight}} = & \ 10 \cdot \mathcal{L}_{\text{normal}} + 
10 \cdot \mathcal{L}_{\text{src\_HDRI}} + 
0.2 \cdot \mathcal{L}_{\text{src\_diff}} \\
& - \mkern-18mu + \ 0.2 \cdot \mathcal{L}_{\text{albedo}} + 
0.2 \cdot \mathcal{L}_{\text{PBR}} + 
0.2 \cdot \mathcal{L}_{\text{Neural}} \\
& - \mkern-18mu + \mathcal{L}_{\text{vgg}_\text{src\_diff}} + 
\mathcal{L}_{\text{vgg}_\text{albedo}} + 
\mathcal{L}_{\text{vgg}_\text{PBR}} + 
\mathcal{L}_{\text{vgg}_\text{Neural}} \\
& - \mkern-18mu + \ \mathcal{L}_{\text{adv}_\text{PBR}} + 
\mathcal{L}_{\text{adv}_\text{Neural}} + 
\ 0.2 \cdot \mathcal{L}_{\text{spec}_\text{Neural}}.
\end{aligned}
\end{equation}

We empirically determined the  weighting coefficients.

%% file: sec/4_pretraining.tex
\section{Multi-Masked Autoencoder Pre-training}

% Relighting challenges models to accurately perceive and manipulate the image features such as structure, color, and texture.

We introduce the Multi-Masked Autoencoder (MMAE), a self-supervised pre-training framework designed to enhance feature representations in relighting models.
It aims to improve output quality without relying on additional, costly light stage data.
Building upon the MAE framework~\cite{he2022masked}, MMAE capitalizes on the inherent learning of crucial image features like structure, color, and texture, which are essential for relighting tasks.
However, adapting MAE to our specific needs poses several non-trivial challenges.
Firstly, MAE is primarily designed for vision transformers~\cite{dosovitskiy2020image}, while our focus is on a UNet, a convolution-based architecture.  This convolutional structure, with its hierarchical nature and aggressive pooling, is known to simplify the MAE task, necessitating careful adaptation~\cite{woo2023convnext}.
Further, the hyperparameters of MAE, particularly the fixed mask size and ratio, are also specific to vision transformers. These factors could introduce biases during training and hinder the model to recognize image features at various scales. 
% Moreover, MAE is optimized for visual recognition downstream tasks, where the decoder is often discarded after pre-training. In contrast, our task, which involves visual generation, requires the full utilization of both the encoder and decoder. 
Moreover, MAE relies solely on masked region reconstruction loss, limiting the model to understand the global coherence of the reconstructed region in relation to its visible context.
% there is a fundamental different in architecture: MAE is designed for vision transformer, whereas our focus is on a UNet, which is convolution-based. The convolution-based networks inherently possess a hierarchical structure, which simplifies the task of MAE due to aggressive pooling compared to vision transformer~\cite{woo2023convnext}.
% Furthermore, MAE is tailored for visual recognition tasks, optimizing the encoding of visible contexts by dropping the decoder after pre-training. In contrast, our application is visual generation, which necessitates the use of both encoder and decoder, without omitting any component.
% Additionally, the hyperparameters, particularly the fixed mask size and ratio, are specifically tied to their vision transformer architecture~\cite{dosovitskiy2020image}.
% These constraints can result in biases during training pair reconstruction, potentially causing the model to overlook image features at varying scales.
% Moreover, relying solely on masked region reconstruction loss hinders model to learn the global coherence of the reconstructed region in relation to its visible context.

\begin{figure}[!t]
  \includegraphics[width=\linewidth]{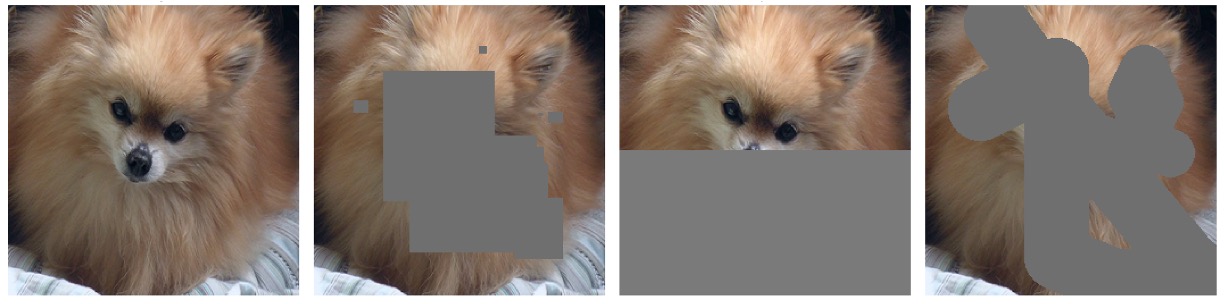}
  \vspace{-6mm}
  % \captionsetup{font=footnotesize}
  \caption{
  \textbf{Dynamic Masking Strategies.} We have generalized the MAE masks to include overlapping patches of varying sizes, as well as outpainting and free-form masks.
  }
  \label{fig:mmae}
  \vspace{-6mm}
\end{figure}

To address these challenges effectively, we have developed two key strategies within the MMAE framework:

\vspace{1mm}
\noindent\textbf{Dynamic Masking.} MMAE eliminates two key hyperparameters, mask size and ratio, by introducing a variety of mask types to generalize the MAE. These types, which include overlapping patches of various sizes, outpainting masks~\cite{teterwak2019boundless}, and free-form masks~\cite{liu2018image} (see Fig.\ref{fig:mmae}), each contribute to the model's versatility.
% Detailed mask construction strategies are provided in the supplementary material.
With the ability to handle challenging masked regions, MMAE achieves a more comprehensive understanding of image properties.

\noindent\textbf{Generative Target.} In addition to its structural advancements, MMAE incorporates a new loss function strategy. 
We have adopted perceptual~\cite{johnson2016perceptual} and adversarial losses~\cite{isola2017image}, along with the original reconstruction loss. 
As a result, MMAE is equipped not only to reconstruct missing image parts but also to ensure synthesis capabilities and their seamless integration with the original context.
In practice, the weights for the three losses are equally set.
\vspace{1mm}

We pre-train the entire UNet architecture using MMAE, and, unlike MAE, we retain the decoder and fine-tune the entire model on relighting ground truths.

% accurate

% As we are using the UNet-based convolutianal networks, MMAE can naturally addresses these limitations by introducing various mask types, including random stroke-typed masks, overlapping patches of varying sizes, and outpainting masks.
% By breaking free from the constraints of fixed-size, non-overlapping patches , this variety enables a more comprehensive grasp of image properties under different lighting conditions, crucial for realistically modeling light dynamics. 

% Masked autoencoder
% fixed mask ratio 0.75 or 0.6
% fixed mask type (fixed-sized patches w.o overlaps)
% l1/l2 reconstruction loss

% \subsection{Approach}
% \paragraph{Dynamic Masking.}
% 1) brush
% 2) varying-sized patches w. overlaps
% 3) outpaint
% \paragraph{Objectives.}
% l1 + vgg + gan loss

%% file: sec/5_data.tex
\section{Data} % and Implementation Details}
% \noindent\textbf{Data Preparation.}
% In line with standard protocol~\cite{sun2019single,nestmeyer2020learning,pandey2021total} 
We constructed the \textbf{OLAT} (One Light at a Time) dataset using a light stage~\cite{debevec2000acquiring,wenger2005performance} equipped with 137 programmable LED lights and 7 frontal-viewing cameras. Our dataset comprises images of 287 subjects, with each subject being captured in up to 15 different poses, resulting in a total of 29,705 OLAT sequences.
We sourced \textbf{HDRI} dataset from several publicly available archives. 
Specifically, we acquired 559 HDRI maps from Polyhaven, 76 from Noah Witchell, 364 from HDRMAPS, 129 from iHDRI, and 34 from eisklotz.
In addition, we incorporated synthetic HDRIs created using the method proposed in~\cite{mei2023lightpainter}. 
During training, HDRIs are randomly selected with equal probability from either real-world or synthetic collections.

We produced training pairs by projecting the sampled source and target lighting maps onto the reflectance fields of the OLAT images~\cite{debevec2000acquiring}. 
To derive the ground truth intrinsics, we applied the photometric stereo method~\cite{woodham1980photometric} and obtained normal and albedo maps. 

%% file: sec/6_experiments.tex
\section{Experiments}

This section details our experimental results.
We begin with a comparative evaluation of our method against state-of-the-art approaches using the OLAT dataset.
We also employ images from the FFHQ--test~\cite{karras2019style} for user studies.
For qualitative analysis, we utilize copyright-free portrait images from Pexels~\cite{pexels}.
Additionally, we conduct ablation studies to validate the efficacy of our pre-training framework and 
architectural design choice.
Subsequently, we detail the additional features and conclude by discussing its limitations.
Our evaluation uses the OLAT test set, comprising 35 subjects and 11 lighting environments, ensuring no overlap with the train set. 

\vspace{1mm}
\noindent\textbf{Evaluation metrics.}
We employ several key metrics for evaluating the prediction accuracy; Mean Absolute Error (\textbf{MAE}), Mean Squared Error (\textbf{MSE}), Structural Similarity Index Measure (\textbf{SSIM}) and Learned Perceptual Image Patch Similarity (\textbf{LPIPS}). While these metrics offer valuable quantitative insights, they do not fully capture the subtleties of visual quality enhancement. Therefore, we emphasize the importance of qualitative evaluations to gain a comprehensive understanding of model performance.

\vspace{1mm}
\noindent\textbf{Baselines.} 
We compared our approach with three state-of-the-art baselines:
Single Image Portrait Relighting (\textbf{SIPR})~\cite{sun2019single}, which uses a single neural network for relighting; Total Relight (\textbf{TR})~\cite{pandey2021total}, employing multiple neural networks that incorporate the Phong reflectance model; and \textbf{Lumos}~\cite{yeh2022learning}, a TR adaptation for synthetic datasets. 
Due to the lack of publicly available code or model from these methods, we either integrated their techniques into our framework or requested the original authors to process our inputs with their models and share the results.

% Specifically, for quantitative analysis, we implemented SIPR and TR within our training and testing framework. This allowed us to assess the impact of network configurations and underlying physics principles on performance. For qualitative comparison, we evaluated our relighting results against SIPR, TR, and Lumos, facilitated by the original authors who applied their models to the same inputs and provided their outputs.
%for comparison.

\begin{table}
\centering
\tablestyle{3pt}{1.2}
\begin{tabular}{l cccc}
\shline
\multicolumn{1}{l}{}        & MAE $\downarrow$ & MSE $\downarrow$ & SSIM $\uparrow$ & LPIPS $\downarrow$ \\
\cline{2-5}
SIPR~\cite{sun2019single}   &0.1715 &0.0748 &0.8432 &3.648 \\
TR~\cite{pandey2021total}   &0.1643 &0.0658 &0.8465 &3.425 \\
% \hline
Ours                        &0.1023 &0.0275 &0.9002 &2.137 \\
\gr
Ours (w. MMAE)              &0.0933 &0.0235 &0.9051 &2.059 \\
\shline
\end{tabular}
\caption{\textbf{Quantitative Evaluation} on the OLAT test set.}
\label{tab:quantitative_eval}
\vspace{-1em}
\end{table}

\begin{table}
\centering
\tablestyle{5pt}{1.2}
\begin{tabular}{l ccg}
\shline
\multicolumn{1}{l}{}        & Lumos~\cite{yeh2022learning} & TR~\cite{pandey2021total} & Ours \\
\cline{2-4}
Consistent Lighting         & 0.0478  &  0.1852  & 0.7671  \\
Facial Details              & 0.2022  &  0.2602  & 0.5376 \\
Similar Identity            & 0.1741  &  0.2440  & 0.5819 \\  
\shline
\end{tabular}
\caption{\textbf{User Study} on the FFHQ test set. }
\label{tab:userstudy}
\vspace{-3mm}
\end{table}

\begin{figure}[!t]
  \includegraphics[width=\linewidth]{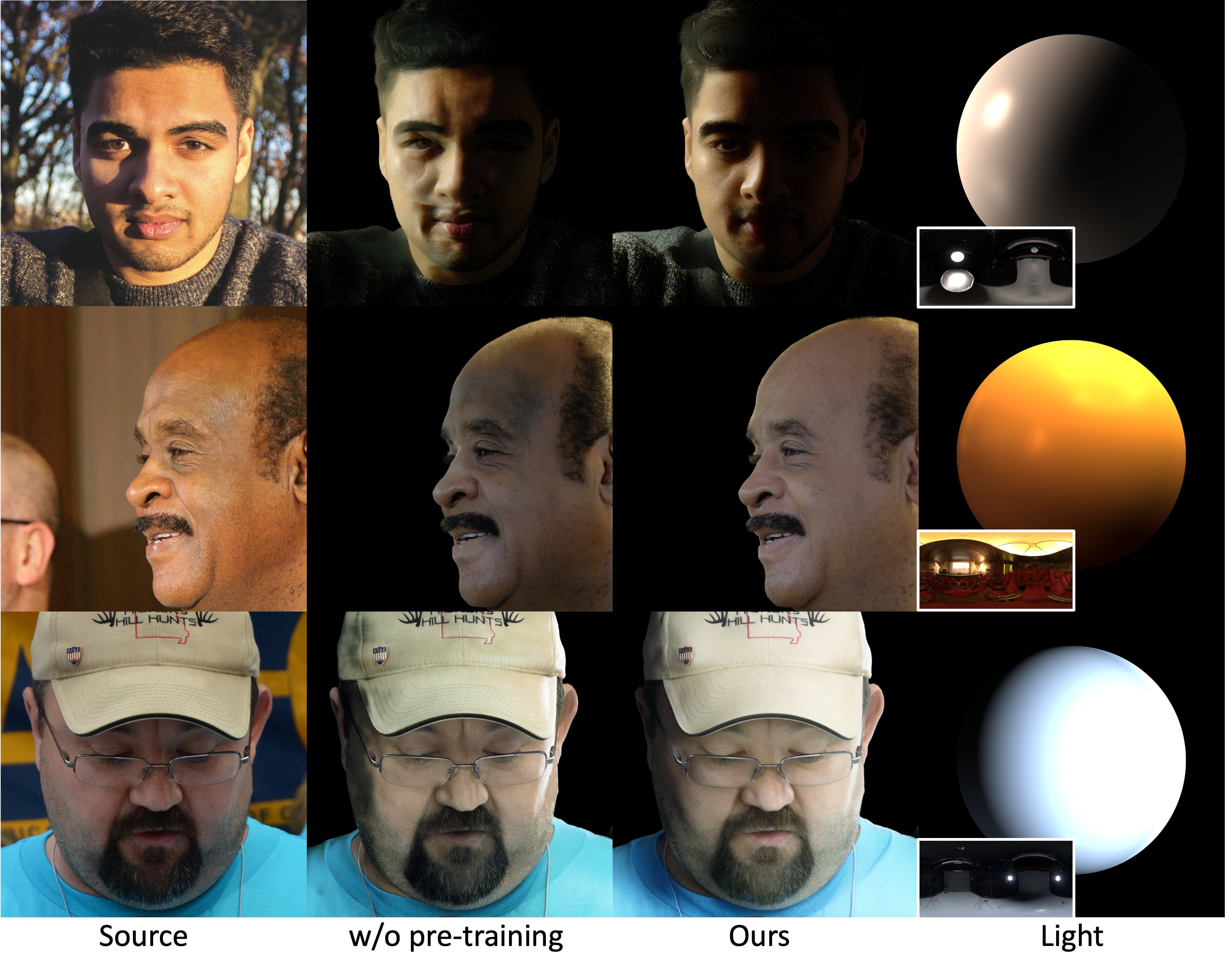}
  \vspace{-8mm}
  % \captionsetup{font=footnotesize}
  \caption{\textbf{Impact of Pre-training.} 
  The fine details such as specular highlights, skin tones, and shadows are notably improved.
  }
  \label{fig:abl_pretrain}
  \vspace{-3mm}
\end{figure}

\begin{figure}[!t]
  \includegraphics[width=\linewidth]{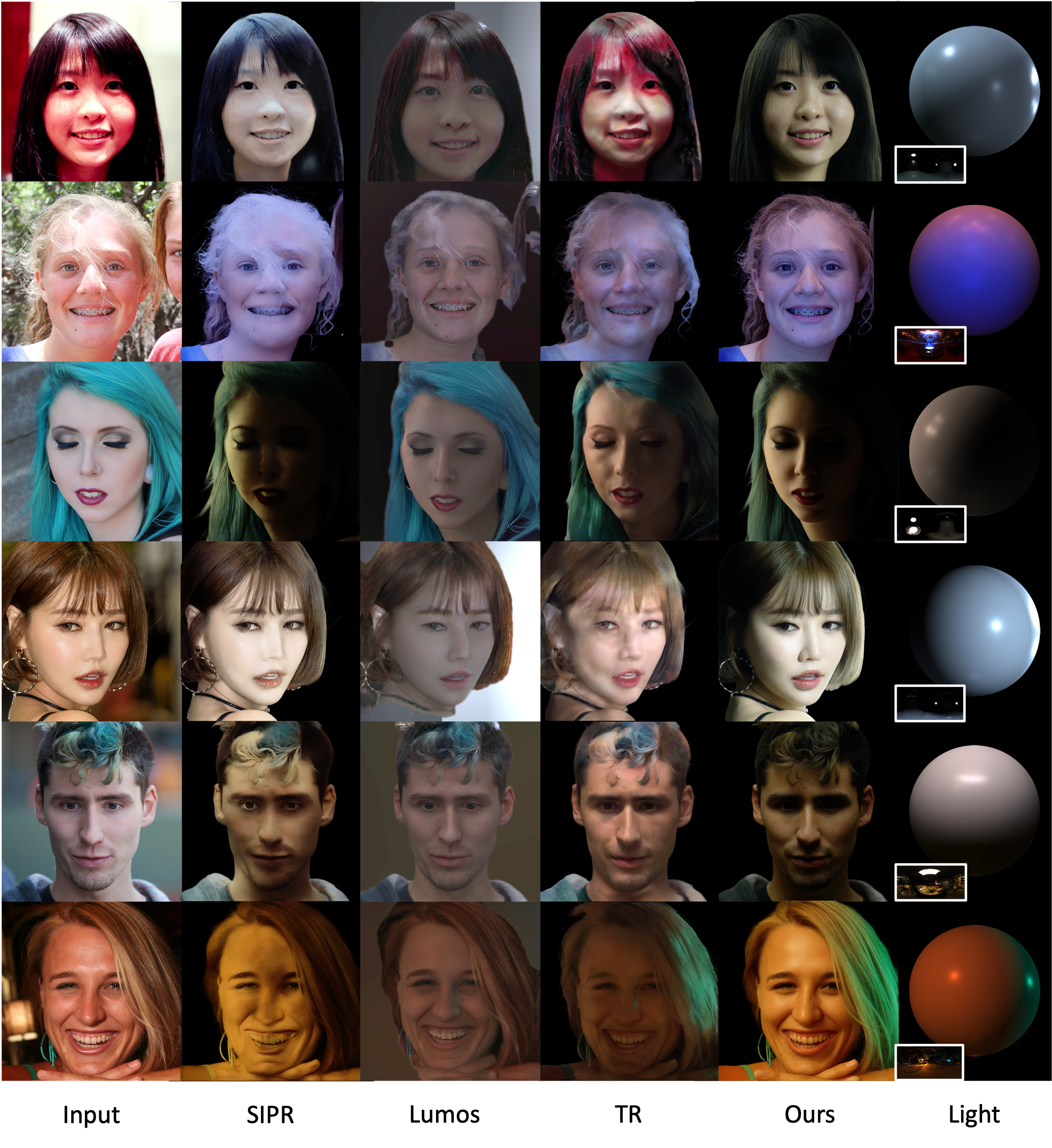}
  \vspace{-6mm}
  % \captionsetup{font=footnotesize}
  \caption{\textbf{Qualitative Comparison} on the Pexels images~\cite{pexels}. 
  %Comparing the proposed approach with state-of-the-art relighitng methods (SIPR, Lumos, and TR) using 
  }
  \label{fig:sota_compare}
  \vspace{-4mm}
\end{figure}

% \vspace{1mm}
\noindent\textbf{Quantitative Comparisons.}
The results in Table.~\ref{tab:quantitative_eval} shows our method outperforming SIPR and TR baselines, demonstrating the significance of incorporating advanced rendering physics and reflectance models. 
The transition from SIPR to TR emphasizes the value of physics-based design, while the shift from TR to our approach underscores the importance of transitioning from the empirical Phong model to the more accurate Cook-Torrance model. Additionally, pre-training contributes to further enhancements, as evidenced by the improved image details, depicted in Fig~\ref{fig:abl_pretrain}.

\vspace{2mm}
\noindent\textbf{Qualitative Comparisons.}
Our relighting method exhibits several key advantages over previous approaches, as showcased in Fig.~\ref{fig:sota_compare}.
It effectively harmonizes light direction and softness, avoiding harsh highlights and inaccurate lighting that are commonly observed in other methods.
A notable strength of our approach lies in its ability to capture high-frequency details like specular highlights and hard shadows.
Additionally, as shown in the second row, it preserves facial details and identity, ensuring high fidelity to the subject's original features and mitigating common distortions seen in previous approaches.
Moreover, our approach excels in handling skin tones, producing natural and accurate results under various lighting conditions. This is clearly demonstrated in the fourth row, where our method contrasts sharply with the over-saturated or pale tones from previous methods.
Finally, the nuanced treatment of hair is highlighted in the sixth row, where our approach maintains luster and detail, unlike the flattened effect typical in other methods.
More qualitative results are available in our supplementary video demonstration.

\vspace{2mm}
\noindent\textbf{User Study.}
We conducted a human subjective test to evaluate the visual quality of relighitng results, summarized in Table.~\ref{tab:userstudy}.
In each test case, workers were presented with an input image and an environment map. 
They were asked to compare the relighting results from three methods--Ours, Lumos, and TR--based on three criteria: 1) consistency of lighting with the environment map, 2) preservation of facial details, and 3) maintenance of the original identity. To ensure unbiased evaluations, the order of the methods presented was randomized.
To aid in understanding the concept of consistent lighting, relit balls were displayed alongside the images. The study included a total of 256 images, consisting of 32 portraits each illuminated with 8 different HDRIs.
Each worker was tasked with selecting the best image for each specific criterion, randomly assessing 30 samples. A total of 47 workers participated in the study. The results indicate a strong preference for our results over the baseline methods across all evaluated metrics.

% \begin{table}
% \centering
% \tablestyle{3pt}{1.2}
% \begin{tabular}{l cccc}
% \shline
% \multicolumn{1}{l}{} & MAE $\downarrow$ & MSE $\downarrow$ & SSIM $\uparrow$ & LPIPS $\downarrow$ \\
% \cline{2-5}
% MAE~\cite{he2022masked}    &0.0952 &0.0242 &0.9007 &2.096 \\
% \gr
% MMAE                       &0.0933 &0.0235 &0.9051 &2.059 \\
% \hline
% Albedo                     &0.1053 &0.0295 &0.8985 &2.197 \\
% \gr
% Diff Render                &0.1023 &0.0275 &0.9002 &2.137 \\
% \shline
% \end{tabular}
% \caption{\textbf{Ablation Studies} on the OLAT test set.}
% \label{tab:quantitative_eval}
% \end{table}

\begin{table}
\centering
\tablestyle{1pt}{1.2}
\begin{tabular}{lccccc}
\shline
&\multicolumn{1}{c}{Method} & MAE $\downarrow$ & MSE $\downarrow$ & SSIM $\uparrow$ & LPIPS $\downarrow$ \\
\cline{3-6}
\multirow{2}{*}{Pre-training}
& MAE~\cite{he2022masked} & 0.0952 & 0.0242 & 0.9007 & 2.096 \\
& \cellcolor{gray!10}MMAE &  \cellcolor{gray!10}0.0933 & \cellcolor{gray!10}0.0235 & \cellcolor{gray!10}0.9051 & \cellcolor{gray!10}2.059 \\
\hline
\multirow{2}{*}{DiffuseNet}
& Albedo & 0.1053 & 0.0295 & 0.8985 & 2.197 \\
& \cellcolor{gray!10}Diff Render  & \cellcolor{gray!10}0.1023 & \cellcolor{gray!10}0.0275 & \cellcolor{gray!10}0.9002 & \cellcolor{gray!10}2.137  \\
\shline
\end{tabular}
\caption{\textbf{Ablation Studies} on the OLAT test set.}
\label{tab:ablation_study}
\vspace{-3mm}
\end{table}

\begin{figure}[!t]
  \includegraphics[width=\linewidth]{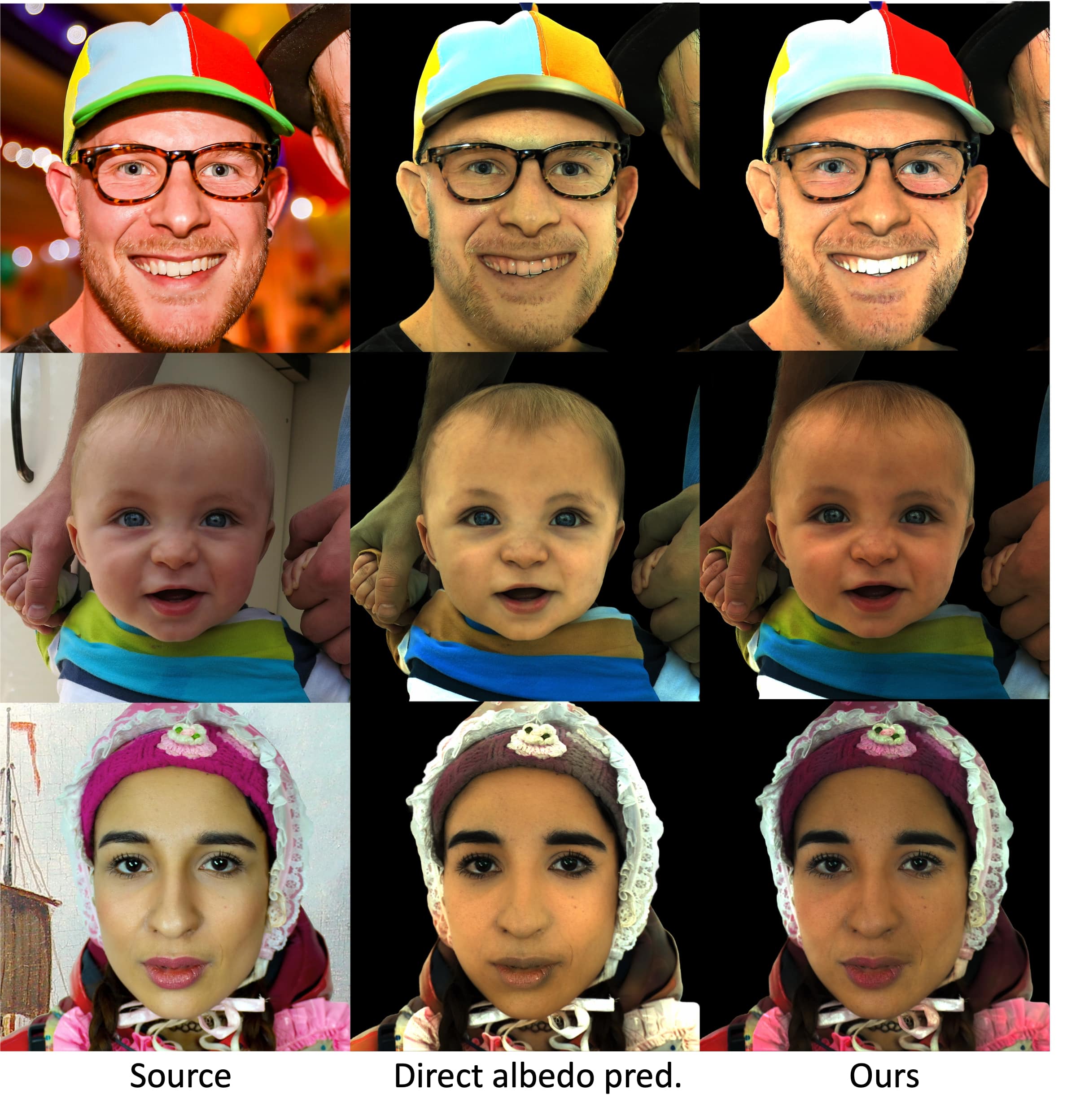}
  \vspace{-8mm}
  % \captionsetup{font=footnotesize}
  \caption{\textbf{Ablation on DiffuseNet.}
  Our approach successfully infers the albedo on various surfaces (skin, teeth, and accessories).}
  \label{fig:abl_albedo}
  \vspace{-5mm}
\end{figure}

\noindent\textbf{Ablation Studies.}
We analyze our two major design choices in Table.~\ref{tab:ablation_study}: the MMAE pre-training framework and DiffuseNet.
The MMAE, which integrates dynamic masking with generative objectives, outperforms MAE.
This superiority is mainly due to the incorporation of challenging masks and global coherence objectives, enabling the model to learn richer features during pre-training.
Furthermore, our method of predicting diffuse render demonstrates superiority over direct albedo prediction.
Firstly, we see it simplifies the learning process, as predicting diffuse render is more closely related to the original image. 
Secondly, our approach effectively distinguishes between the influences of illumination (diffuse shading) and surface properties (diffuse render). This distinction is crucial for accurately modeling the intrinsic color of surfaces, as it enables independent and precise evaluation of each element (see Eqn.~\ref{eq:ours_albedo}).
In contrast, methods that predict albedo directly often struggle to differentiate between these factors, leading to significant inaccuracies in color constancy, as shown in Fig. ~\ref{fig:abl_albedo}.

\begin{figure}[!t]
  \includegraphics[width=\linewidth]{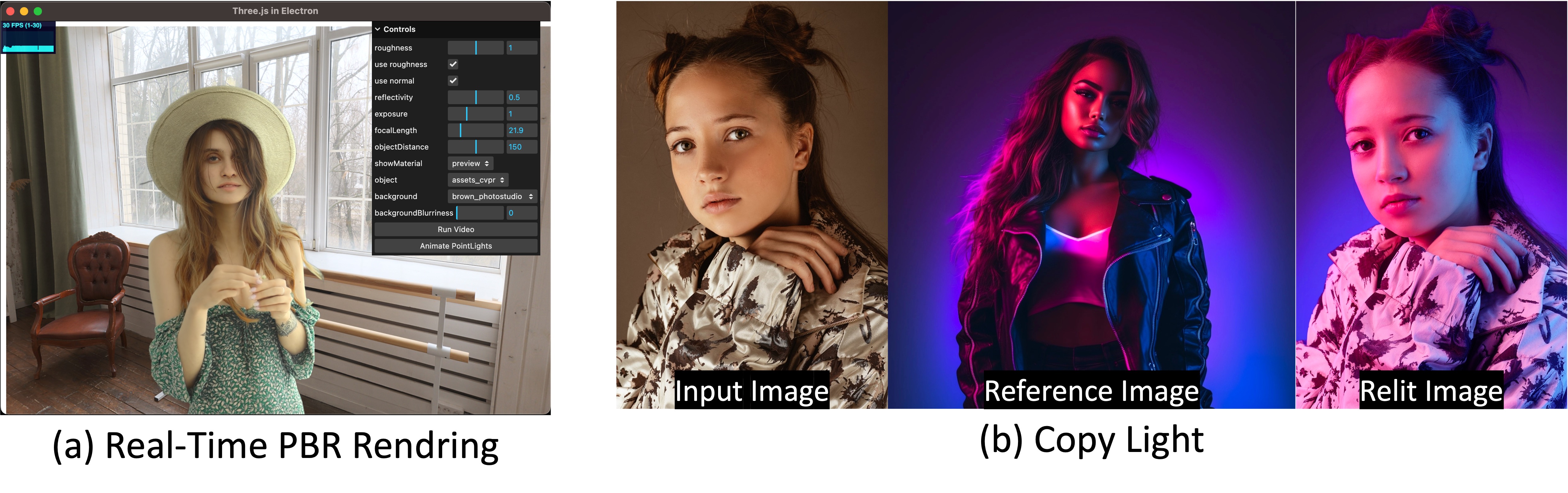}
  \vspace{-8mm}
  % \captionsetup{font=footnotesize}
  \caption{\textbf{Applications.} 
  We showcase additional features of SwitchLight, powered by the diverse intrinsics features.
  }
  \label{fig:application}
  \vspace{-3mm}
\end{figure}

\begin{figure}[!t]
  \includegraphics[width=\linewidth]{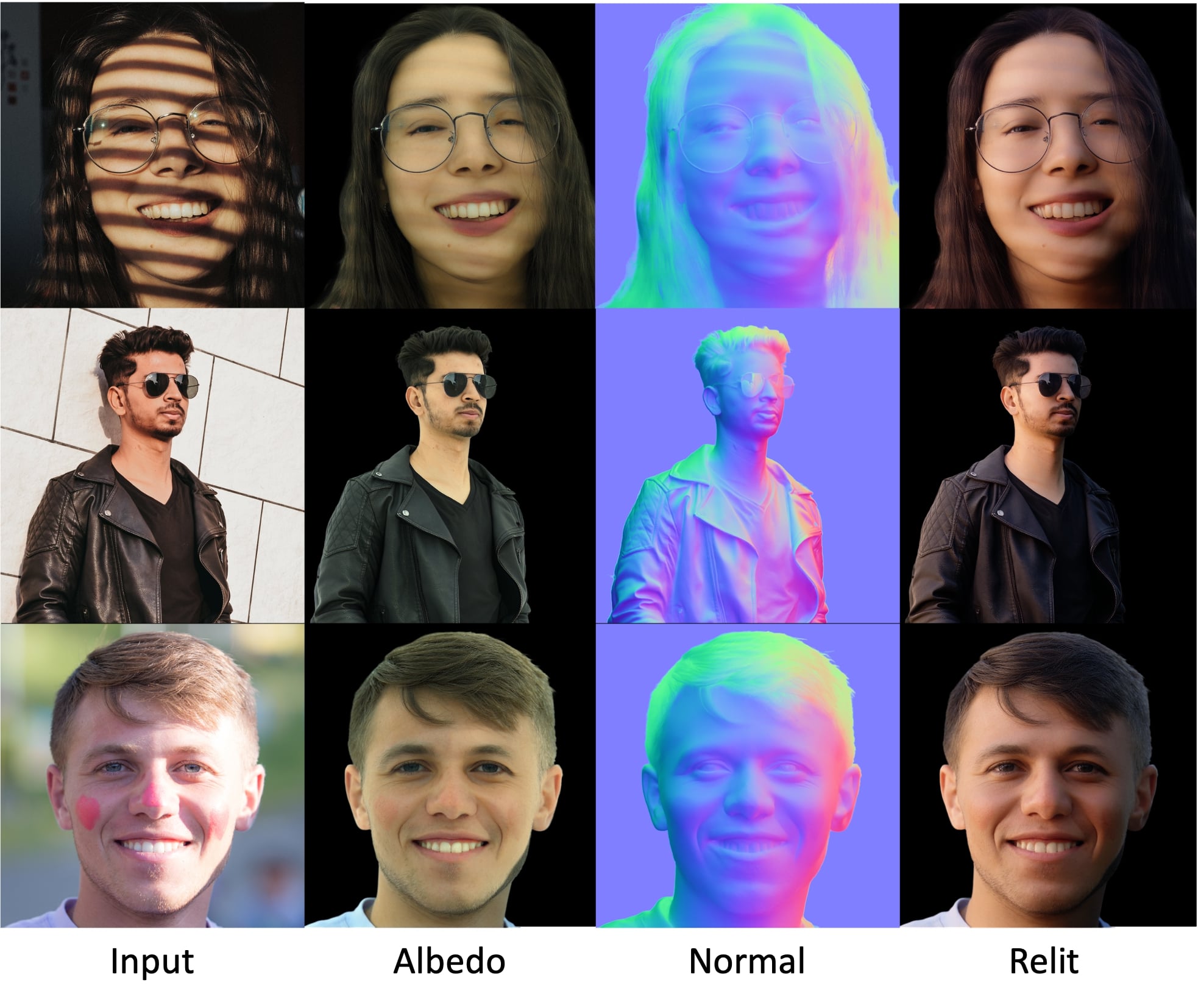}
  \vspace{-8mm}
  % \captionsetup{font=footnotesize}
  \caption{\textbf{Limitations.} 
  The model faces challenges in removing strong shadows, misinterpreting reflective surfaces like sunglasses, and inaccurately predicting albedo for face paint.
  %, impacting relit image fidelity.
  % leading to residual shadow artifacts in albedo and relit images. Reflective surfaces like sunglasses are misinterpreted as opaque, affecting specularity elimination. Inaccurate albedo prediction for face paint also impacts relit image fidelity.
  }
  \vspace{-3mm}
  \label{fig:limitation}
  \vspace{-2mm}
\end{figure}

% pre-training
% -mae to ours
% -l1 to l1/vgg/gan

\vspace{2mm}
\noindent\textbf{Applications.}
We present two applications using predicted intrinsics in Fig.~\ref{fig:application}.
First, real-time PBR via Cook-Torrance components in Three.js graphics library.
Second, switching the lighting environment between the source and reference images.
Further details are in the supplementary video.

\vspace{2mm}
\noindent\textbf{Limitations.}
We identified a few failure cases in Fig.~\ref{fig:limitation}.
First, our model struggles with neutralizing strong shadows, which leads to inaccurate facial geometry and residual shadow artifacts in both albedo and relit images. 
Incorporating shadow augmentation techniques~\cite{zhang2020portrait,futschik2023controllable} during training could mitigate this issue.
Second, the model incorrectly interprets reflective surfaces, such as sunglasses, as opaque objects in the normal image. This error prevents the model from properly removing reflective highlights in the albedo and relit images.
Lastly, the model inaccurately predicts the albedo for face paint.
Implementing a semantic mask~\cite{yeh2022learning} to distinguish different semantic regions separately from the skin could help resolving these issues.
%It would allow the model to recognize and process different semantic regions, like accessories or face paint, separately from the skin. This targeted approach could enhance the accuracy of intrinsic predictions.

%% file: sec/7_conclusion.tex
\vspace{-2mm}
\section{Conclusion}

We introduce SwitchLight, an architecture based on Cook-Torrance rendering physics, enhanced with a self-supervised pre-training framework.
This co-designed approach significantly outperforms previous models.
Our future plans include scaling the current model beyond images to encompass video and 3D data. 
We hope our proposal serve as a new foundational model for relighting tasks.

%% file: sec/X_suppl.tex
% \clearpage
% \setcounter{page}{1}
% \maketitlesupplementary

\appendix
\section*{\Large{Appendix}}

\section{Implementation Details}
\noindent\textbf{Training}
Our method incorporates both pre-training, spanning for 100 epochs, and fine-tuning phases lasting 50 epochs.
In the pre-training stage, we utilized the ImageNet dataset, using a batch size of 1024 images, each with a resolution of 256 \(\times\) 256 pixels.
We employed the Adam optimizer; 10k linear warm-up schedule followed a fixed learning rate of \(1e^{-4}\).
In the fine-tuning stage, we switched to the OLAT datset, with a batch size reduced to 8 images, each at a resolution of \(512 \times 512\) pixels.
The Adam optimizer is used with a fixed learning rate of \(1e^{-4}\).
The entire training process takes one week to converge using 32 NVIDIA A6000 GPUs.

We pre-train a single U-Net architecture during this process. In the subsequent fine-tuning stage, the weights from this pre-trained model are transferred to multiple U-Nets - NormalNet, DiffuseNet, SpecularNet, and RenderNet. In contrast, IllumNet, which does not follow the U-Net architecture, is initialized with random weights. To ensure compatibility with the varying input channels of each network, we modify the weights as necessary. For example, weights pre-trained for RGB channels are copied and adapted to fit networks with 6 or 9 channels.

\vspace{3mm}
\noindent\textbf{Data}
To generate the relighting training pairs, we randomly select each image from the OLAT dataset. Two randomly chosen HDRI lighting environment maps are then projected onto these images to form a training pair. The images undergo processing in linear space. For managing the dynamic range effectively, we apply logarithmic normalization using the \( \text{log}(1+x) \) function. 
% This approach has been observed to stabilize the training process.

\vspace{3mm}
\noindent\textbf{Architecture}
SwitchLight employs a UNet-based architecture, consistently applied across its \textbf{Normal} Net, \textbf{Diffuse} Net, \textbf{Specular} Net, and \textbf{Render} Net. This approach is inspired by recent advancements in diffusion-based models~\cite{dhariwal2021diffusion}.
Unlike standard diffusion methods, we omit the temporal embedding layer.
The architecture is characterized by several hyperparameters:
the number of input channels, a base channel, and channel multipliers that determine the channel count at each stage. Each downsampling stage features two residual blocks, with attention mechanisms integrated at certain resolutions.
The key hyperparameters and their corresponding values are summarized in Table.~\ref{tab:arch_spec}.

\begin{table}[t!]
\centering
\small % This command sets the font size to small
\tablestyle{2pt}{0.8}
\begin{tabular}{lcccc}
\toprule
 & Normal Net & Diffuse Net & Specular Net & Render Net \\
\midrule
In ch         & 3    & 6    & 9   & 9    \\
Base ch          & 64   & 64   & 64   & 64   \\
Ch mults    & [1,1,2,2,4,4] & [1,1,2,2,4,4] & [1,1,2,2,4,4] & [1,1,2,2,4,4] \\
Num res         & 2    & 2    & 2    & 2    \\
Head ch         & 64   & 64   & 64   & 64   \\
Att res  & [8,16,32] & [8,16,32] & [8,16,32] & [8,16,32] \\
Out ch        & 3    & 3    & 2    & 3    \\
\bottomrule
\end{tabular}
\caption{\textbf{Network Architecture Parameters.}
This table outlines the key hyperparameters and their corresponding values; initial input channels (In ch), base channels (Base ch), and channel multipliers (Ch mults) that set the stage-specific channel counts. 
It also indicates the number of residual blocks per stage (Num res), the number of channels per head (Head ch), the stages where attention mechanisms are applied based on feature resolution (Att res), and the final output channels (Out ch).
}
\vspace{-3mm}
\label{tab:arch_spec}
\end{table}

%illumnet
\textbf{IllumNet} is composed of two projection layers, one for transforming the Phong lobe features and another for image features, with the latter using normal bottleneck features as a compact form of image representation. Following this, a cross-attention layer is employed, wherein the Phong lobe serves as the query and the image features function as both key and value. Finally, an output layer generates the final convolved source HDRI.

%discriminator
The \textbf{Discriminator} network is utilized during both pre-training and fine-tuning stages, maintaining the same architectural design, although the weights are not shared between these stages. This network is composed of a series of residual blocks, each containing two 3\(\times\)3 convolution layers, interspersed with Leaky ReLU activations. 
The number of filters progressively increases across these layers: 64, 128, 256, and 512. Correspondingly, as the channel filter count increases, the resolution of the features decreases, and finally, the network compresses its output  with a 3x3 convolution into a single channel, yielding a probability value.

Regarding the activation functions across different networks:
NormalNet processes its outputs through \(\ell_2\) normalization, ensuring they are unit normal vectors.
IllumNet, DiffuseNet, and RenderNet utilize a softplus activation (with \(\beta = 20\)) to generate non-negative pixel values.
SpecularNet employs a sigmoid activation fuction,  ensuring that both the roughness parameter and Fresnel reflectance values fall within a range of 0 to 1.

\begin{figure}[!t]
  \includegraphics[width=\linewidth]{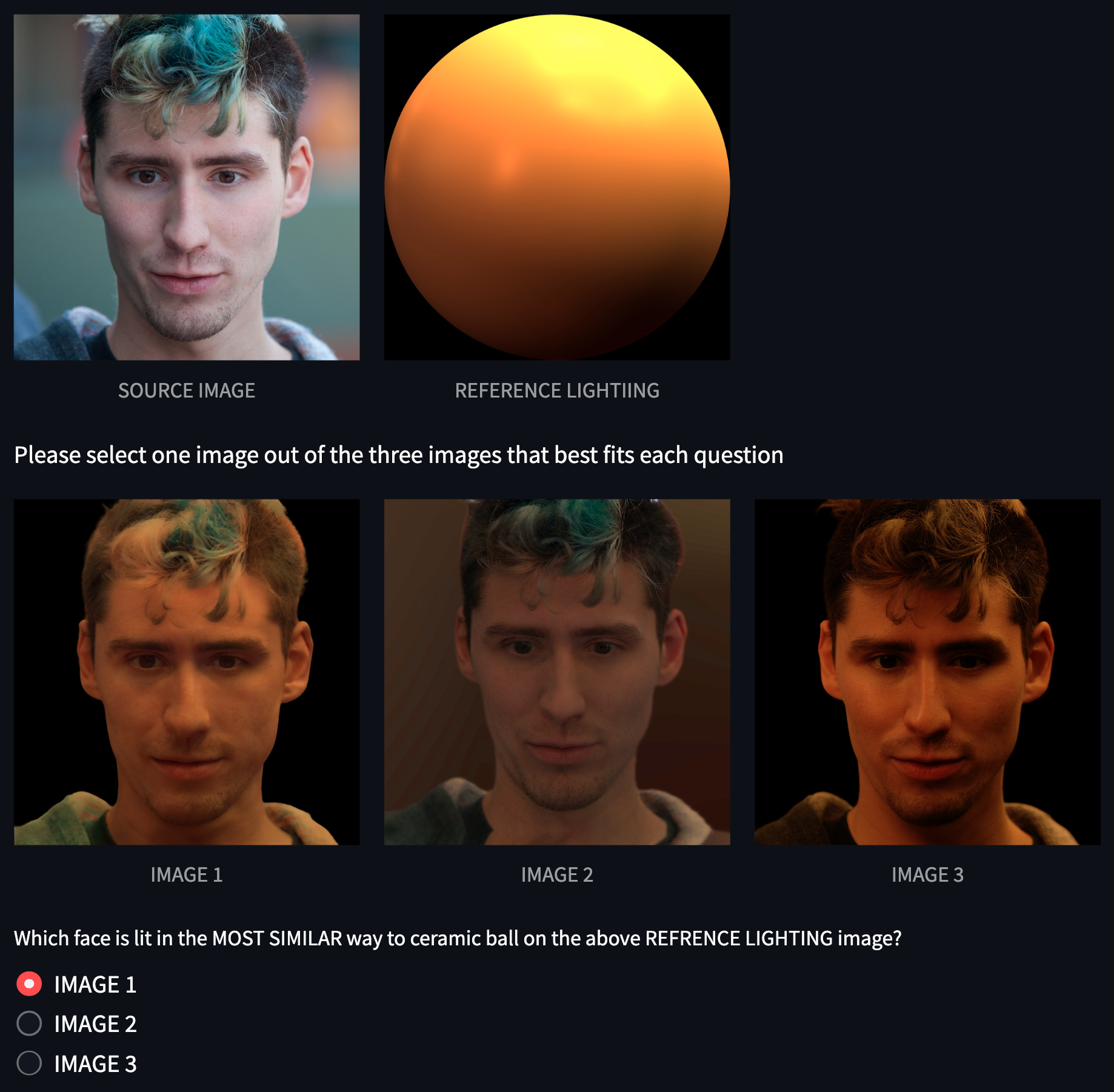}
  \caption{\textbf{User Study Interface} comparing relighting results with prior approaches, focusing on consistency in lighting, preservation of facial details, and retention of original identity.
  }
  \label{fig:interface}
  \vspace{-3mm}
\end{figure}

\section{User Study Interface}
% Our user study interface is demonstrated as follows: We present participants with an input image alongside a diffused ball illuminated by the target environment map. The main focus is on comparing our relighting result with the baselines, as shown in Fig.~\ref{fig:interface}. The evaluation criteria are: 1) Which image exhibits more consistent lighting? 2) Which maintains better facial details? 3) Which retains the original identity more effectively? The comparison aims to identify which image not only aligns more closely with the lighting of the diffused ball but also excels in preserving facial details and the original identity.

% We ensured unbiased evaluations by randomizing the presentation order and using relit balls to facilitate understanding of lighting consistency. Each worker randomly assessed 30 samples from a batch of 256.
% This set included 32 portraits, each illuminated under 8 different lighting conditions. The images were sourced from the FFHQ dataset~\cite{karras2019style}.

Our user study interface is outlined as follows: Participants are shown an input image next to a diffused ball under the target environment map lighting. The primary objective is to compare our relighting results with baseline methods, as depicted in Fig.~\ref{fig:interface}. Evaluation focuses on three criteria: 1) Consistency of lighting, 2) Preservation of facial details, and 3) Retention of the original identity. This comparison aims to determine which image best matches the lighting of the diffused ball while also maintaining facial details and original identity.
To ensure unbiased evaluations, we randomized the order of presentation. Participants evaluated 30 random samples from a set of 256. This dataset included 32 portraits from the FFHQ dataset~\cite{karras2019style}, each illuminated under eight distinct lighting conditions.

\section{Video Demonstration}
We present a detailed video demonstration of our SwitchLight framework. Initially, we use real-world videos from Pexels~\cite{pexels} to showcase its robust generalizability and practicality. Then, for state-of-the-art comparisons, we utilize the FFHQ dataset~\cite{karras2019style} to demonstrate its advanced relighting capabilities over previous methods. The presentation includes several key components:

\begin{enumerate}\itemsep1em
    \item \textbf{De-rendering:} This stage demonstrates the extraction of normal, albedo, roughness, and reflectivity attributes from any given image, a process known as inverse rendering.
    \item \textbf{Neural Relighting:} Leveraging these intrinsic properties, our system adeptly relights images to align with a new, specified target lighting environment.
    \item \textbf{Real-time Physically Based Rendering (PBR):} Utilizing the Three.js framework and integrating derived intrinsic properties with the Cook-Torrance reflectance model, we facilitate real-time rendering. This enables achieving 30 fps on a MacBook Pro with an Apple M1 chip (8-core CPU and 8-core GPU) and 16 GB of RAM.
    \item \textbf{Copy Light:} Leveraging SwitchLight's ability to predict lighting conditions of a given input image, we explore an intriguing application. This process involves two images, a source and a reference. We first extract their intrinsic surface attributes and lighting conditions. Then, by combining the source intrinsic attributes with the reference lighting condition, we generate a new, relit image. In this image, the source foreground remains unchanged, but its lighting is altered to match that of the reference image. 
    \item \textbf{State-of-the-Art Comparisons:} We benchmark our framework against leading methods, specifically Total Relight~\cite{pandey2021total} and Lumos~\cite{yeh2022learning}, to highlight substantial performance improvements over these approaches.
\end{enumerate}

\section{Additional Qualitative Results}
\begin{figure*}[!t]
  \centering
  \includegraphics[width=\textwidth,height=0.99\textheight,keepaspectratio]{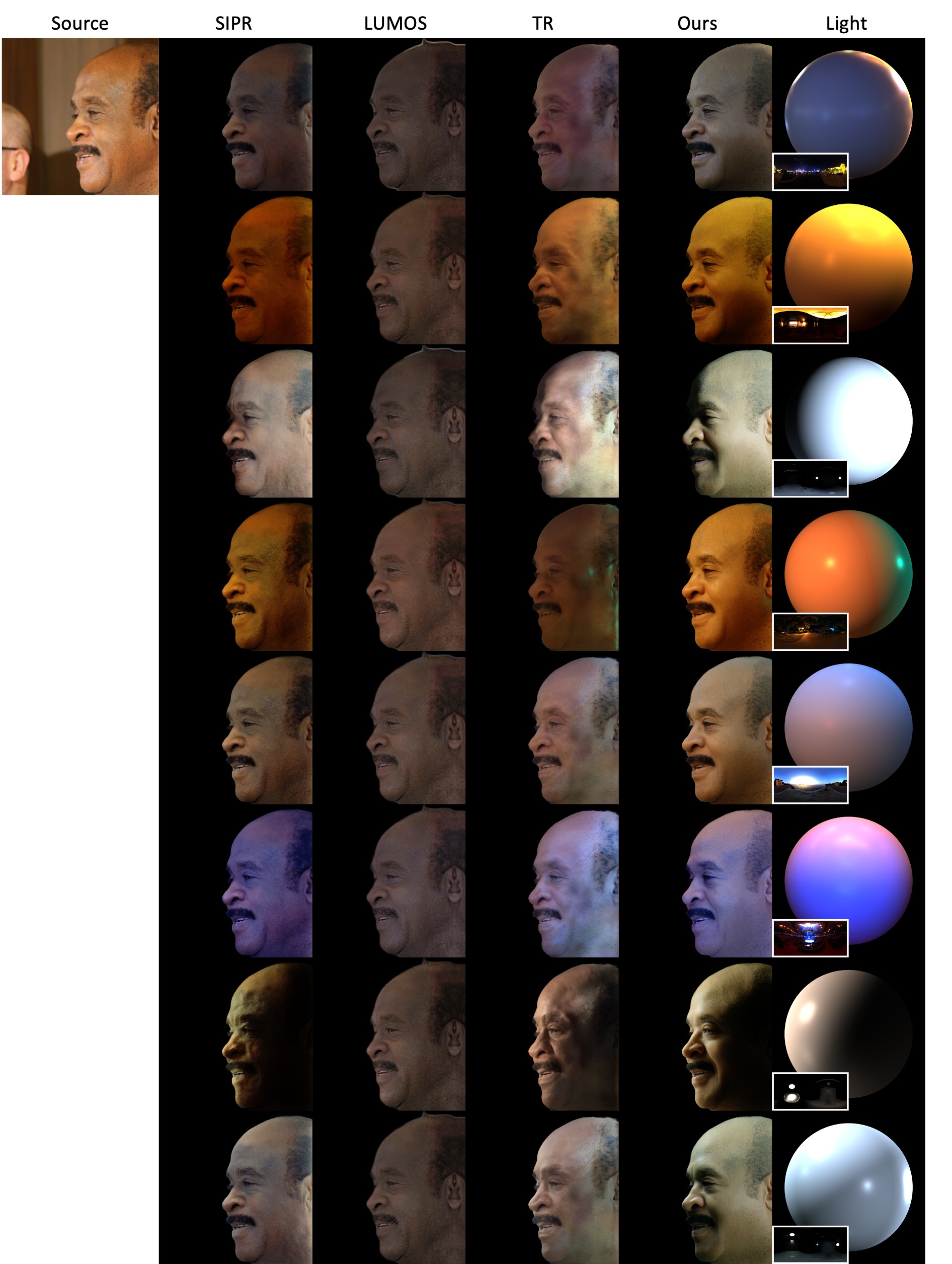}
  \caption{\textbf{Qualitative Comparisons} with state-of-the-art approaches.
  }
  \label{fig:qual_1}
\end{figure*}

\begin{figure*}[!t]
  \centering
  \includegraphics[width=\textwidth,height=0.99\textheight,keepaspectratio]{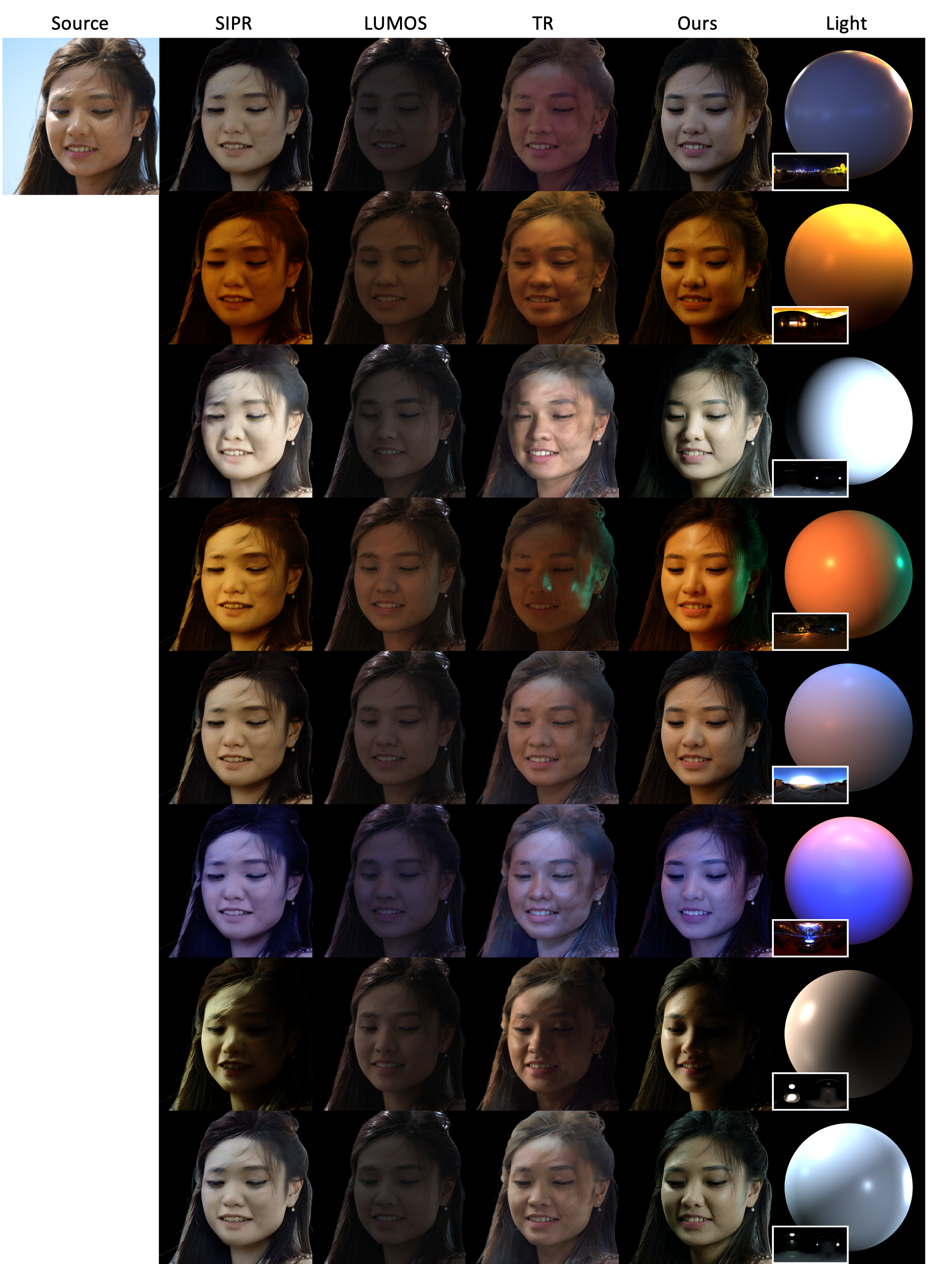}
  \caption{\textbf{Qualitative Comparisons} with state-of-the-art approaches.
  }
  \label{fig:qual_2}
\end{figure*}

\begin{figure*}[!t]
  \centering
  \includegraphics[width=\textwidth,height=0.99\textheight,keepaspectratio]{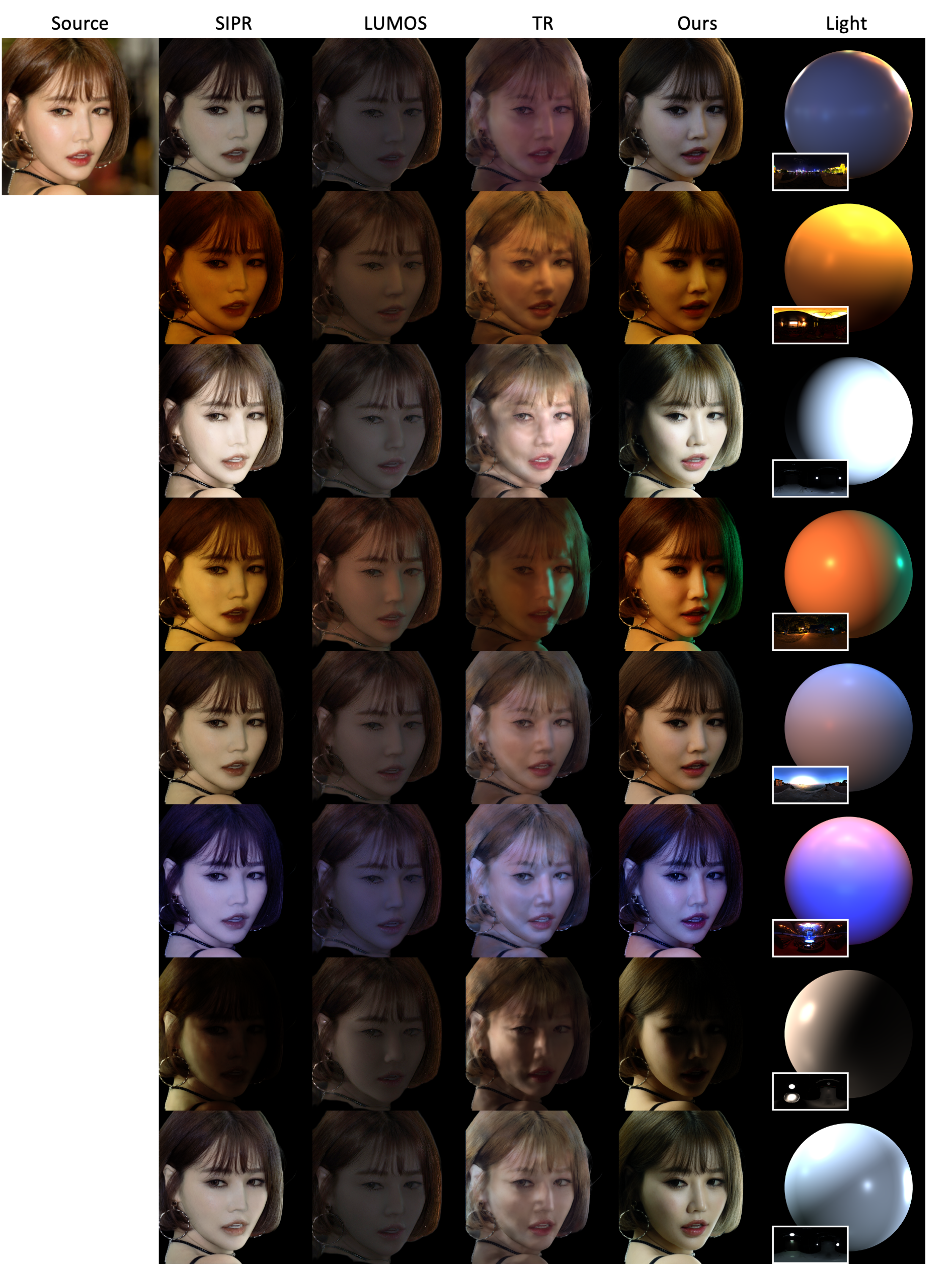}
  \caption{\textbf{Qualitative Comparisons} with state-of-the-art approaches.
  }
  \label{fig:qual_3}
\end{figure*}

\begin{figure*}[!t]
  \centering
  \includegraphics[width=\textwidth,height=0.99\textheight,keepaspectratio]{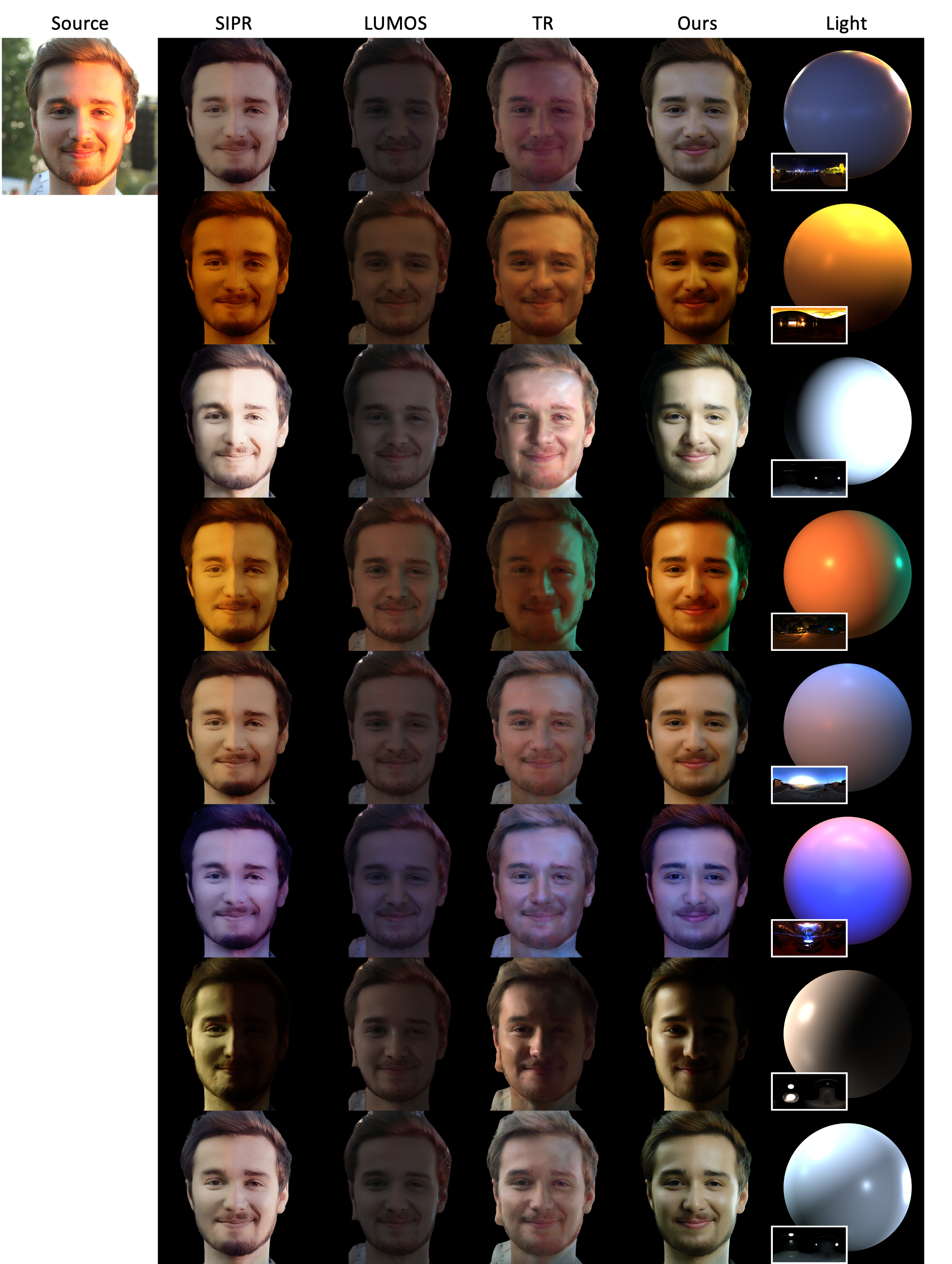}
  \caption{\textbf{Qualitative Comparisons} with state-of-the-art approaches.
  }
  \label{fig:qual_4}
\end{figure*}

\begin{figure*}[!t]
  \centering
  \includegraphics[width=\textwidth,height=0.99\textheight,keepaspectratio]{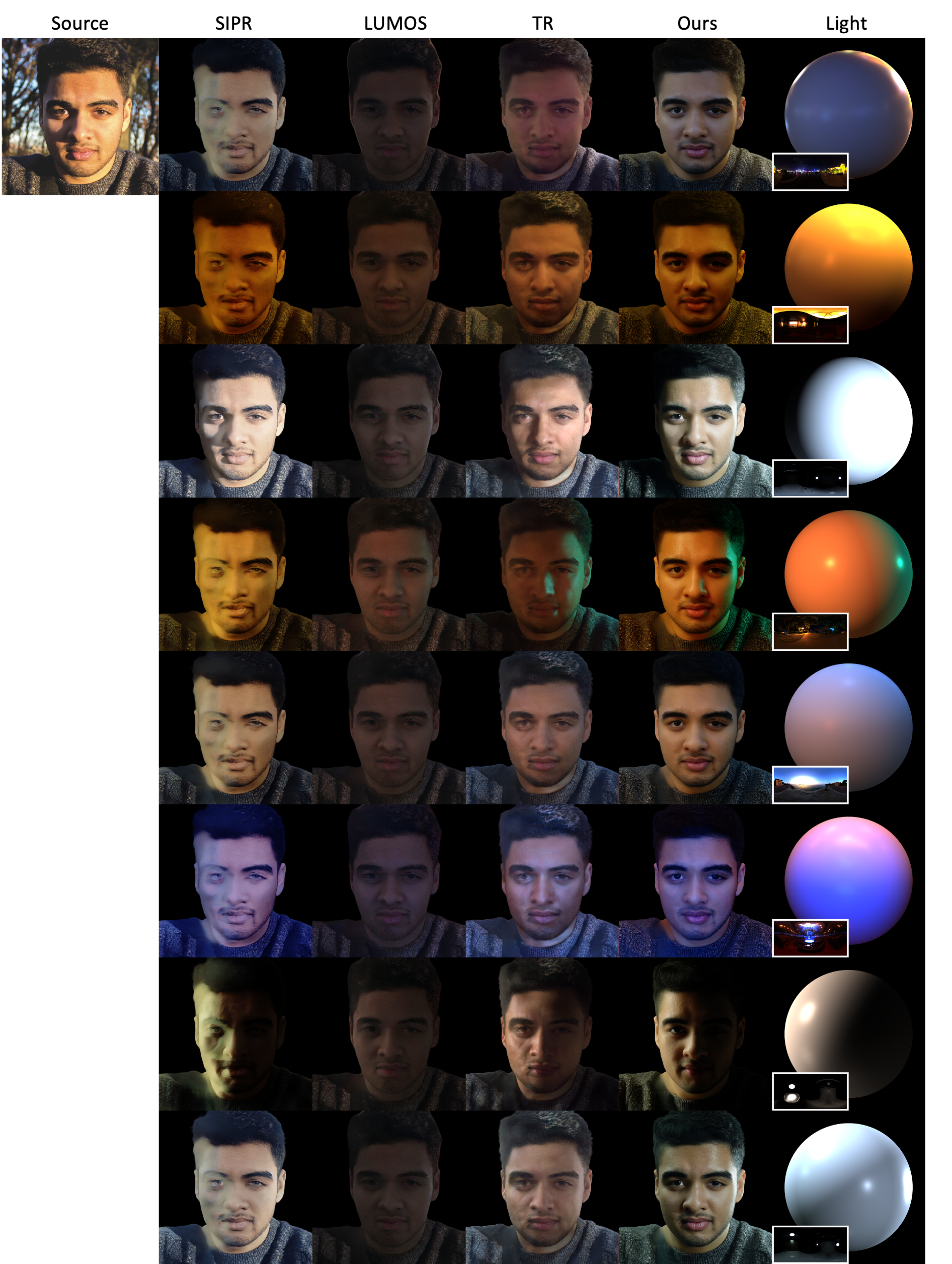}
  \caption{\textbf{Qualitative Comparisons} with state-of-the-art approaches.
  }
  \label{fig:qual_5}
\end{figure*}

Further qualitative results are provided in Fig.\ref{fig:qual_1}, \ref{fig:qual_2}, \ref{fig:qual_3}, \ref{fig:qual_4}, and \ref{fig:qual_5}. 
Each figure illustrates the relighting of a source image in eight distinct target lighting environments. 
In these figures, our approach is benchmarked against prior state-of-the-art methods, namely SIPR~\cite{sun2019single}, Lumos~\cite{yeh2022learning}, and TR~\cite{pandey2021total}, utilizing images from Pexels~\cite{pexels}. This comparison is enabled by the original authors who applied their models to identical inputs and provided their respective outputs.

We can clearly observe that our method demonstrates notable efficacy in achieving consistent lighting, maintaining softness and high-frequency detail. Additionally, it effectively manages specular highlights and hard shadows, while meticulously preserving facial details, identity, skin tones, and hair texture.

